\documentclass[sigconf, nonacm]{acmart}
\usepackage{xcolor}
\usepackage{svg}
\usepackage{amsmath,amsfonts}
\usepackage{algorithmic}
\usepackage{graphicx}
\usepackage{textcomp}
\usepackage{xcolor}
\usepackage{graphicx}
\usepackage[ruled, vlined] {algorithm2e}
\usepackage{amsmath}
\usepackage{epsfig}
\usepackage{tcolorbox}
\usepackage{graphicx,amsmath,subfigure,stfloats, url}
\usepackage{multirow}
\usepackage{listings}
\usepackage{balance}
\usepackage{soul}
\usepackage{xcolor}
\usepackage{hyperref}
\usepackage[normalem]{ulem}
 \usepackage[T1]{fontenc}
\usepackage{aecompl}
\newcommand\vldbpagestyle{plain} 
\newcommand{\eat}[1]{}
\newcommand{\stitle}[1]{\noindent{\bf #1}}
\newcommand{\bl}[1]{#1}

\begin{document}
%\title{Contextualized soft token Prompts and Information Augmentation: Tackling Low-Resource GEM Challenges}
%\title{Contextualized soft token Prompts and Information Augmentation: Tackling Low-Resource GEM Challenges}
%\title{Extracting and Inserting: Contextualized soft token Enhanced and Information Augmented Prompts for Tackling Low-Resource GEM Challenges}

\title{\bl{APrompt4EM: Augmented Prompt Tuning for Generalized Entity Matching}}
%%
%% The "author" command and its associated commands are used to define the authors and their affiliations.

\author{Yikuan Xia}
\affiliation{Key Laboratory of High Confidence Software Technologies, CS, Peking University, Beijing, China}
\email{wfl00014@pku.edu.cn}
\authornote{Equal Contribution}

\author{Jiazun Chen}
\affiliation{
\institution{Key Laboratory of High Confidence Software Technologies, CS, Peking University, Beijing, China}
}
\email{chenjiazun@stu.pku.edu.cn}
\authornotemark[1]

\author{Xinchi Li}
\affiliation{
\institution{Big data and Artificial Intelligence Institute of the China Telecom Research Institute, Beijing, China}
}
\email{lixc6@chinatelecom.cn}

\author{Jun Gao}
\affiliation{Key Laboratory of High Confidence Software Technologies, CS, Peking University, Beijing, China}
\email{gaojun@pku.edu.cn}
\authornote{Corresponding Author} 
%%
%% The abstract is a short summary of the work to be presented in the
%% article.
\begin{abstract}
\bl{Generalized Entity Matching (GEM), which aims at judging whether two records represented in different formats refer to the same real-world entity, is an essential task in data management. The prompt tuning paradigm for pre-trained language models (PLMs), including the recent PromptEM model, effectively addresses the challenges of low-resource GEM in practical applications, offering a robust solution when labeled data is scarce. However, existing prompt tuning models for GEM face the challenges of prompt design and information gap. This paper introduces an augmented prompt tuning framework for the challenges, which consists of two main improvements. The first is an \textbf{augmented} contextualized soft token-based prompt tuning method that extracts a guiding soft token benefit for the PLMs' prompt tuning, and the second is a cost-effective information \textbf{augmentation} strategy leveraging large language models (LLMs). Our approach performs well on the low-resource GEM challenges. Extensive experiments show promising advancements of our basic model without information augmentation over existing methods based on moderate-size PLMs (average $5.24\%$+), and our model with information augmentation achieves comparable performance compared with fine-tuned LLMs, using less than $14\%$ of the API fee.}
\end{abstract}

\maketitle

%%% do not modify the following VLDB block %%
%%% VLDB block start %%%
\pagestyle{\vldbpagestyle}

\section{Introduction}

\label{sec:intro}
With the growth of data volume and the increasing demand for high-quality data, addressing the issue of Entity Matching (EM) has become an urgent need in many applications, i.e., data integration~\cite{doan2012principles} and cleaning~\cite{chu2016data}. The term EM refers to the problem of identifying whether two data instances correspond to the same real-world entity. In the classical setting of the EM problem, we aim to identify all pairs of entity records across two sets that satisfy a matching condition. The well-studied standard EM pipeline consists of two parts~\cite{papadakis2020blocking,konda2018magellan}, the blocking part and the matching part. The blocking part involves efficiently selecting a few entity pairs worthy of comparison, and the matching part is about a fine-grained comparison of the selected entity pairs. In this paper, we focus on the matching part of the EM task. 

\bl{Generalized Entity Matching (GEM) expands upon traditional EM by accommodating a broader variety of data formats beyond just structured relational tables~\cite{wang2021machamp}. Conventional EM relies on identical or aligned schemas and seeks similarity between pairs of identical attributes~\cite{mudgal2018deep}. In contrast, GEM aims to handle a broader range of data types, including relational, semi-structured, and textual data types. Such a flexibility requirement makes GEM closer to real-world applications where entities are represented in diverse formats. As EM is a special case of GEM, and GEM can deal with a wider range of real-world tasks, we will deal with the GEM problem in this paper.} Besides, existing main-stream EM methods rely on labeled training data heavily~\cite{li2020deep,mudgal2018deep}, and \bl{obtaining labeled data for GEM is often a challenge because manual labeling of data is labor-intensive, time-consuming, and requires domain expertise, making the process costly and slow, especially considering the diverse and dynamic nature of data sources~\cite{wang2012crowder,gokhale2014corleone}.} How to build a GEM system leveraging limited labels is a realistic challenge. So, in this paper, we mainly focus on the problem of GEM in a low-resource setting.

\begin{figure}
    \centering
    \includegraphics[width=0.5\textwidth]{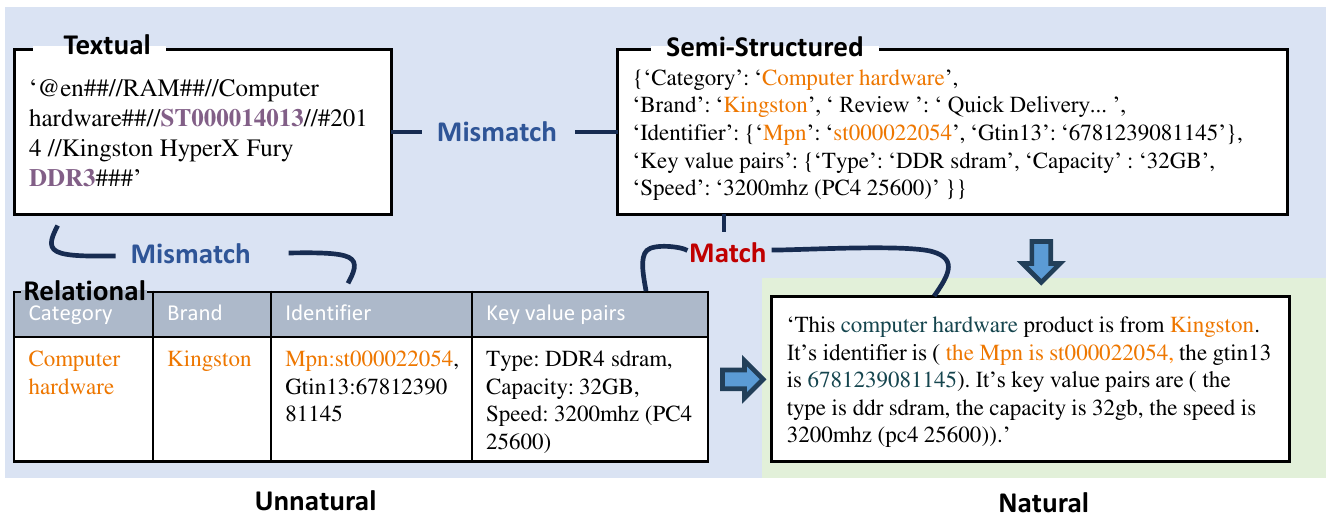}
    \caption{Different data schemas in GEM and unifying them using natural language texts. The entity presented by semi-structured and relational data is a typical RAM product in 2022, while the entity presented in text is a typical RAM product in 2014.}
    \label{fig:intro}
\end{figure}

\bl{The GEM problem, as a generalized version of the EM problem, not only inherits its core challenges, the diverse \textbf{semantic} representations and the \textbf{noisy} data~\cite{barlaug2021neural}, but also introduces a novel difficulty: \textbf{flexible} data format.} These difficulties are further deepened, given the scarcity of labeled resources. For instance, in Fig.~\ref{fig:intro}, the terms `computer hardware' and `IT equipment' from two matching entities have the same semantic meaning (\textbf{semantic}). The key name `Key value pairs' or the review information are not helpful in matching the entities (\textbf{noisy}). One source may present data in a nested JSON format detailing product attributes, while the other could be mere text describing product features (\textbf{flexible}). From a human's perspective, the critical information for discriminating between these two entities, the product identifier (Mpn, Manufacturer Product Number) and the equipment parameters (highlighted in different colors in Fig.~\ref{fig:intro}), is either hidden in the redundant data or missing. As illustrated by such cases, the key challenge lies in how to exploit the useful semantic information from divergent data representations.

Tuning Pre-trained Language Models (PLMs) is the mainstream method of addressing the semantic issue. Previous methods~\cite{li2020deep} adopt the fine-tuning training paradigms to adapt the pre-existing knowledge within PLMs to the specific matching scenario. To address the low-resource label issue, prompt-based methods, known for their resource efficiency, lead the extensively trained PLMs to extract relevant insights with fewer labels~\cite{liu2023pre}. PromptEM ~\cite{wang2022promptem} is the representative method of prompt tuning for EM, which uses structural serialization~\cite{li2020deep} and the P-tuning model~\cite{liu2023gpt} as its prompt. Given the distinct advantage of prompt-based methods under limited resource conditions, our research pivots around this technique. 

However, existing prompt strategies are  faced with limitations:

\stitle{Limitation I:} Issues related to prompt design. The previous methods do not adequately address the format flexibility and noisy data. As we have mentioned, PromptEM's prompt design consists of structured serialization and P-tuning soft tokens~\cite{wang2022promptem}. The currently used prompt of PromptEM poses two issues. Firstly, the PLM's input of PromptEM, following the footsteps of Ditto~\cite{li2020deep}, serializes entities using identifiers like [COL] and [VAL]. This serialization does not align with the natural language patterns encountered during PLMs' pre-training, as structural or semi-structural data is limited in PLMs' training data~\cite{trabelsi2022strubert}. Secondly, in the GEM landscape, directly serializing the entities results in mismatches in field types and data formats, which, combined with the noisy data issue, can diffuse the model's focus and lead to inferior outcomes. We have to adjust the prompt to fit the distribution, which the LM is familiar with. The P-tuning-style soft token method PromptEM uses  searches for better prompts for the task, which partly alleviates this issue. However, the prompt template is unified for one dataset, and does not benefit each entity on purpose. 

\stitle{Limitation II:} Issues related to semantic information gap. The semantic issue is not fully addressed, especially when language models (LMs) are tuned using a few labels. The semantic relationship may be hard for LMs to capture, or the key information is not provided in the entity. Though PLMs may contain information in their training corpus, the prior semantic information of moderate-size PLMs (i.e., BERT~\cite{devlin2018bert}, RoBERTa~\cite{liu2019roberta}, GPT~\cite{radford2018improving,radford2019language}) may not be sufficient for complex domain-specific EM~\cite{ji2023survey}. For example, in Fig.~\ref{fig:intro}, we might expect our model to tell the two products (presented in semi-structured, textual format) through the information that the mainstream DRAM product in 2014 is not likely to have 64GB of storage. We find out that if we input `The mainstream storage of a 2014 RAM is [MASK] GB.' to a BERT model~\cite{devlin2018bert}, the output with top-2 probability is 32 and 64, indicating that the PLM is unaware of the mainstream RAM equipment parameters of each year. This may deteriorate the matching performance. Conversely, a supplement based on objective facts can relieve the problem.

In this paper, we introduce a prompt tuning framework, APrompt\-4EM, which makes substantial improvements to address these two challenges. To address the first limitation, we tailor a template generation method for the matching scenario, converting structured or unstructured data into natural language patterns. The natural language prompt is more aligned with the PLMs' training context, which boosts the model's performance. \eat{Incorporating a `mask-in-the-middle' approach,} Further, we augment the prompt with a soft token mechanism. Contrary to the task-specific soft tokens in previous works~\cite{liu2023gpt}, the contextualized soft token utilizes a neural network to extract essential portions from the noisy data to form instance-specific soft tokens, which guides the model's focus and enables the model to converge faster.

To address the second limitation, we emphasize the importance of information augmentation. We propose augmenting sufficient information to bridge semantic gaps, especially when the PLM's intrinsic knowledge falls short. Since large language models (LLMs) have shown commendable performance in understanding common knowledge~\cite{wei2022emergent,openai2023gpt4,petroni2019language}, we propose a method to gain insights from LLMs, augmenting the essential information for EM. \eat{Our information-augmented prompt can learn from the hidden knowledge of LLMs, achieving comparable results (LLMs can achieve good results on EM~\cite{peeters2023using,narayan2022can}) while being more cost-effective.} We also propose an uncertainty-based augmentation strategy to reduce the token fees.

Our contributions are summarized as follows:
\begin{itemize}
\item \stitle{Augmented Contextualized Soft Token Based Prompts for Entity Matching.} We propose an augmented prompt tuning model for GEM, in which the natural language templates and contextualized soft token module, that extract essential information for each entity, can guide the model toward making better decisions. Our prompt-tuning model based on the contextualized soft token shows superior results (an average $\bl{5.24\%}$+ boost) compared with the state-of-the-art methods.
\item \stitle{LLM-supported Information Augmentaion.} We propose augmenting the information to enhance the model's performance. We discuss augmenting information using LLMs to enhance GEM. The cost of API calls is further reduced based on the uncertainty estimation strategy. Our cost-effective information augmentation method can help our base model achieve competitive results compared with a fine-tuned LLM with less than $14\%$ cost.

\end{itemize}

\stitle{Outline.} We first introduce the APrompt4EM framework and its design choices in Sec.~\ref{sec:aprompt4em framework}. We then describe our soft-prompt model and information augmentation strategy in detail in Sec.~\ref{sec:Contextualized} and Sec.~\ref{sec:info}. Finally, we present the experimental results in Sec.~\ref{sec:exp}, the related works in Sec.~\ref{sec:related}, and the conclusions in Sec.~\ref{sec:conclusion}.

 \begin{figure}
    \centering
    \includegraphics[width=\linewidth]{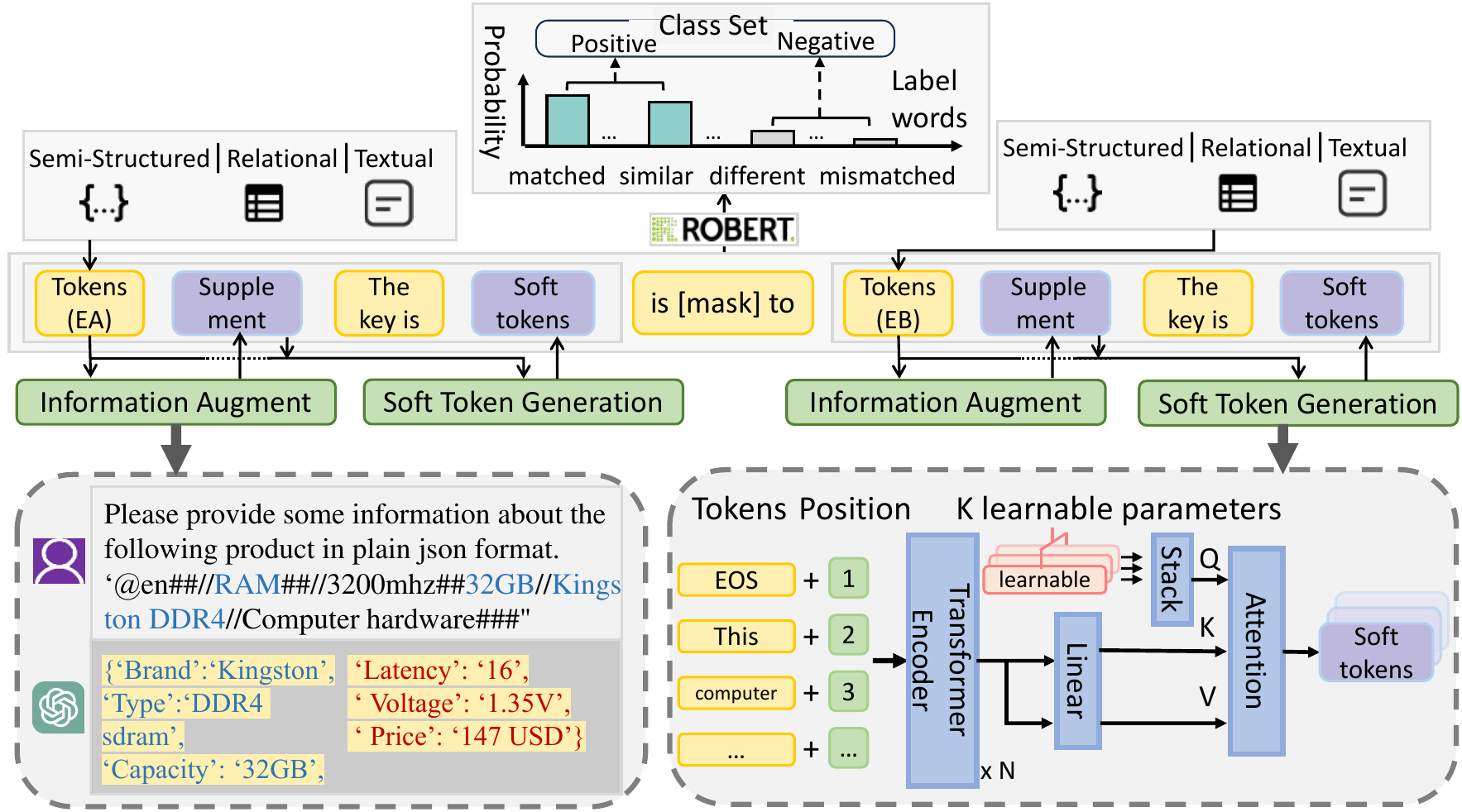}
    \caption{\bl{Illustration of our APrompt4EM framework. }\eat{The left lower part of the figure is the information augmentation module, the right-lower part of the figure is the contextualized soft token module, and the upper part is the natural language prompt module.}}
    \label{fig:enter-label}
\end{figure}

\section{APrompt4EM Framework}
\label{sec:aprompt4em framework}

\bl{In this section, we first discuss the data input for the GEM task. Then we discuss different design choices, and we present the overall framework of our Prompt4EM in Fig.~\ref{fig:enter-label}.}

\stitle{Data input.} This paper, like other works ~\cite{wang2021machamp}, handles three major input data types, including structured tables (relational tables), semi-structured input (JSON format data), and unstructured input (textual data), in a GEM problem.

We formally describe these data, as we need a unified way to handle them, like fitting different inputs into a language model. For structured tables with attribute keys $key_1,....,key_n$, we denote one of its entities $e$ as $e=[\{key_i,val^e_i\},i=1,...,n]$. As for semi-structured input, we consider a nested data structure $S$. To be specific, for an entity $e$, we denote $e=S^e$ as:
\begin{equation*}
    S^e = \left\{
        \begin{aligned}
            &[\{key^e_i, S^e_i\}, & i=1,...,n]\\
            &[S^e_i, & i=1,...,n]
        \end{aligned}
    \right.
\end{equation*}
where $S^e_i$, as the value of a dictionary or the item of a list, is also the defined nested data structure $S$. For unstructured input, we denote one of its entities $e$ as $T^e$, representing its textual content.

Another perspective on the GEM input is the heterogeneity of the data sources, which should be taken into account in module design. For homogeneous data sources, the modules can be designed based on the granularity of the data source. For heterogeneous data sources, the modules have to be designed at the level of the data instance.

\stitle{\bl{Design Choices.}} \bl{As we have stated in Sec.~\ref{sec:intro}, we need to handle three main challenges in GEM, including the semantic issue, the noisy data issue, and the flexible format issue, in one unified method. We can utilize the advancement of language models, extract key features, and convert different data to natural language to handle these three challenges, respectively. Previous works propose different ways of solving the semantic issue. So in the following paragraph, we will make an analysis of our choices and conduct experimental studies for further analysis.}

\bl{To handle the semantic issues, Pre-trained Language Models (PLMs) are introduced to understand the semantic meanings of the entities. There are two kinds of methods to fine-tune PLMs in specific contexts. One is to learn representation for each entity first and then capture the matching rule~\cite{reimers2019sentence}. The other is letting PLMs directly handle the entity pair~\cite{li2020deep}. As the pairwise based method can learn the interaction of features from both entities, it usually achieves better results than the former one~\cite{paganelli2022analyzing}. So we take the pairwise approach as the matching backbone. In addition, to solve GEM in low-resource settings, we follow the prompt tuning paradigm to tune the PLM.}

\bl{We have mentioned in Sec.~\ref{sec:intro} that the semantic issues are not fully addressed, as in some cases, PLMs with moderate-size parameters (e.g., BERT) are insufficient to capture all the semantic information provided in the entity descriptions. One way to bridge the semantic gap is to use LLM as the backbone of LM, and another way is to supplement the semantic information we need for EM. We choose the second way, because although LLM can provide high quality feedback, it is expensive to fine-tune and infer the model. So we take a more economic and practical approach, which is mainly based on fine-tuning a moderate-size PLM, but with enhanced entity information from an LLM.}

\stitle{\bl{Framework of APrompt4EM.}} \bl{APrompt4EM is based on prompt-tuning a PLM (RoBERTa~\cite{liu2019roberta} in this paper) with the input of two entities in natural language format, as illustrated in Fig.~\ref{fig:enter-label} (upper part). The two entities are serialized into a natural language, and they are concatenated and sent to the PLM as a prompt. We obtain the value of [MASK] in the middle of two entities, from which the match or mismatch between two entities is derived.}

\bl{There are two key components in APrompt4EM. As there are noises in the original form and derived form, we introduce the contextualized soft token module (lower-right part in Fig.~\ref{fig:enter-label}) to extract the key features of the entity, through which the impacts of noise data can be alleviated. The soft tokens are generated with the learned parameters for each entity. Another component, the information augmention module (lower-left part in Fig.~\ref{fig:enter-label}), enriches the features of a given entity from a LLM (ChatGPT in our case). We can see that some attributes in yellow can be obtained from LLM, which are missing in the original input.}

\section{Prompt Tuning for GEM}
\label{sec:Contextualized}
In this section, we describe our soft-prompt model design in detail. We first introduce our natural language-based prompt serialization, which serves as a basic discrete prompt. Then, we propose our contextualized soft token prompt model, which is a continuous prompt model. Finally, we describe the training details of the prompt-tuning model.

\subsection{Natural Language Prompt Templates}
\label{sec:natural_prompt}
\stitle{Prompt Template.} Solving the GEM problem under the prompt-based tuning paradigm requires converting the entities to a token sequence using the prompt template. The entities are usually first serialized to a language sequence, and the serialized sequences are then fit into the prompt template for further computation. We follow the prompt template proposed in PromptEM~\cite{wang2022promptem}. Given two entities $e_a,e_b$, the prompt funtion $f_{pt}$ to be filled is defined as:
\begin{equation}
    f_{pt}(e_a,e_b)=serialize(e_a)\ is \ [MASK] \ to \ serialize(e_b).
\end{equation}

\stitle{Previous Structured Entity Serialization.} The serialization function conventions established in Ditto~\cite{li2020deep} can be described as follows. Ditto serializes entities with special tokens. For an entity $e$ as $e=[\{key_i,val^e_i\},i=1,...,n]$ in a structured table, such entity can be serialized as:
\begin{equation*}
    serialize(e)=[COL]\ key_1\ [VAL]\ val^e_1\ ... [COL]\ key_n\ [VAL]\ val^e_n.
\end{equation*}
The $[COL]$ and $[VAL]$ tokens are special tokens to help the model identify the key and value names of the attributes. 

As for semi-structured tables, PromptEM extended the serialization method to semi-structured tables to deal with different input schemas. For nested attributes in the semi-structured tables, the $[COL]$ and $[VAL]$ tokens are recursively added in each nested level, for a nested entity $S^e=[\{key_i^s,S_i^e\}, i=1,..,n]$:
\begin{equation*}
    sl(S^e)=[COL]\ key_1^s\ [VAL]\ sl(S_1^e)\ ...\ [COL]\ key_n^s\ [VAL]\ sl(S_n^e),
\end{equation*}
where $sl$ is the abbreviation of $serialize$. For list objects, the list elements are concatenated into one string to present in the template.

\stitle{\bl{Entity Serialization using Natural Language.}} \bl{Previous works have shown that the prompt templates closer to the LM's data distributions can help the LM retrieve its known knowledge~\cite{jiang2020can}. Although the serialization of Ditto enables the template text to encode the structure information, the use of $[COL]$ and $[VAL]$ tokens is clearly out of the natural language distribution. Therefore, we propose using natural language prompts to help the LM extract information better.}

\bl{For the manual design of the natural language prompt, the basic natural language serializing module for $e=[\{key_i,val^e_i\},i=1,...,n]$ is:}
\begin{equation*}
    \bl{serialize(e)=the\ key_1\ is\ val^e_1,...,the\ key_n\ is\ val^e_n}
\end{equation*}  
\bl{Similar to the processing method of nested attributes in PromptEM, we transform the nested structures into text by recursively applying the basic module. For a nested entity $S^e=[\{key_i^e,S_i^e\}, i=1,..,n]$, we serialize one of its nested attribute value pairs as:}
\begin{equation*}
    \bl{serialize(e)=the\ key_1^e\ is\ sl(S_1^e),...,the\ key_n^e\ is\ sl(S_n^e)},
\end{equation*} 
where $sl$ is the abbreviation of $serialize$.

\bl{Apart from the basic rules, we refer to ChatGPT's concatenation results for entities, and we design different prompts based on various attribute combinations. We use the following prompt to design prompts from ChatGPT:}
\begin{tcolorbox}[colframe=black, colback=white, boxrule=1pt, sharp corners]
\noindent  \textit{\bl{Concatenate the attributes to form a natural language prompt template for `type' entities. Use a declarative sentence in the simple present tense. \\ Example: Attribute: `title', `brand', Prompt Design: This `title' is from `brand'.\\ Please design the prompt template: Attribute: \{attribute\_list\}.}}
\end{tcolorbox}
 \noindent \bl{where `type' represents the dataset scenario, e.g., book/ music.}

\begin{table*}[]
\centering
\caption{Discrete Natural Language Prompt Template Design.}
\resizebox{\textwidth}{!}{%
\begin{tabular}{l|l|l}
\hline
Entity   Schema                                                                                                    & Manual Prompt                                                                                                                                                                                       & Paraphrased   Prompt                                                                                                                                                                                 \\ \hline
\{`title', `author', `venue', `year'\}                                                                             & The `title' is authored by `author' and is published in `venue' in the year `year'.                                                                    & The paper entitled `title' is written by `author'  and is published in `venue'  in  `year'.                                                               \\ \hline
\bl{\{`title', `director', `actors', `year', `rating', `information'\}  }                                                                           &\begin{tabular}[c]{@{}l@{}}  \bl{The `title' is directed by `director', including `actors'.It was released in `year' }\\ \bl{and has received ratings of `ratings'. It includes sl(`information'). }\end{tabular}                                                                 & \begin{tabular}[c]{@{}l@{}}  \bl{The `title' is made  by `director', featuring  `actors'. It debuted  in `year'} \\\bl{and has earned ratings of `ratings'. It encompasses sl(`information'). }\end{tabular}                                   \\ \hline
 \{`category', `brand', `identifiers', `keyvaluepairs', `price'\}      
 & \begin{tabular}[c]{@{}l@{}}This `category' product is from `brand' priced at `price'.  It is identified by `identifiers' \\ and has key value pairs of sl(`keyvaluepairs'). \end{tabular}          & \begin{tabular}[c]{@{}l@{}} This   `category' product is priced by `brand' `price', it is identified by `identifiers' \\ and has key value pairs of sl(`keyvaluepairs') \end{tabular}     \\\hline
\begin{tabular}[c]{@{}l@{}}  \{`name', `postalcode', `latitude', `longitude',\\ `position', `address'\} \end{tabular} 
& \begin{tabular}[c]{@{}l@{}} The `name' is located at `address', with latitude `latitude' and longitude `longitude'. \\ The position is `position', and the postal code is `postalcode'\end{tabular}  
& \begin{tabular}[c]{@{}l@{}} The `name' is located in  the `Address' with the width `Latitude'  and the length `Langitude', \\ the   position is `position' and the postcode is `postalcode'.\end{tabular}  
\\ \hline
\{`title', `brand', `price', `pricecurrency', `description'\}           
& 
The `title' from `brand'. It is priced at `price' `pricecurrency'.  It includes `description'. 
& The `title' from `brand', costs `price' `pricecurrency'.  It includes `description'.         \\ \hline
\{`title', `price', `manufacturter'\}               & The `title' is a product manufactured by `manufacturter' and is priced at `price'.           & The `title' is a product produced by `manufacturer'  and valued at `price'.
\\ \hline
\bl{\{`title', `address', `phone', `category'\} }        & \bl{The `title' is a `category' restaurant located at `address'.  The phone number is `phone'.\   }        &  \bl{The `title' is a `category' restaurant at the `address',  the telephone  number is `phone'.}
\\ \hline
\bl{\{`addr', `city', `phone', `type', `class'\}  }       &\begin{tabular}[c]{@{}l@{}}  \bl{This `type' restaurant   offers a diverse `class' different types of dishes, located  in `city' at `addr'. \\  \bl{The phone number is `phone'.}}\end{tabular} 
& \begin{tabular}[c]{@{}l@{}} \bl{This `type' restaurant   offers a diverse `class' different types of dishes, located  in `city' at `addr'.} \\  \bl{The telephone number is `phone'.}\end{tabular} 
\\ \hline
\begin{tabular}[c]{@{}l@{}}
\bl{\{`song\_name', `artist\_name', `album\_name', `genre',}  \\ \bl{`price',`copyright',`time',`released'\}} \end{tabular}   
&\begin{tabular}[c]{@{}l@{}}   
\bl{The `song\_name' is performed by `artist\_name' and is featured on the album `album\_name'.}\\ \bl{It falls under the genre of `genre'  with a price of `price'. The song is protected by `copyright'}\\ \bl{and has a duration of `time'. It was released on `released'.}  \end{tabular}   
& \begin{tabular}[c]{@{}l@{}} \bl{The `song\_name' is played  by `artist\_name' and is included  on the album `album\_name',}\\ \bl{which falls  under the genre of `genre'  with a price of `price'. The song is protected by `copyright'}\\ \bl{and has a length  of `time'. It was released on `released'.}\end{tabular}   

\\ \hline
\bl{\{`title', `category', `brand', `modelno', `price'\}  }                                         &  \bl{The `title' is a `category' from `brand' with model number `modelno', priced at `price'.}          & 
 \bl{The `title' is a `category' from `brand' with model number `modelno', priced at `price'. }

\\ \hline

\end{tabular}%
}
 
\label{tab:prompt}
\end{table*}

\bl{\stitle{Prompt Template Paraphrasing.} To generate prompt templates closer to the LM's language distribution, we also adopt the paraphrasing approaches proposed in~\cite{jiang2020can} to mine better prompt templates. We use a T5-based~\cite{raffel2020exploring} pre-trained neural translation machine $f_{NMT}$ to first translate manually designed prompts $p$ to another target language (i.e., German) prompt $p_t$ and back translate it to an English prompt $p^{\prime}$. To be specific, for each entity in the training set (of size $n$), we use the beam search strategy to preserve $k_b$ translated $p_t$ and $k_b^2$ back-translated result prompts (e.g., $k_b=3$). We can parse the entity values out of all the $n*k_b^2$ translated result prompts, and keep all the prompt templates if they are successfully parsed. The prompt template with the highest sum of generation probability $p_{NMT}(p_t|p)p_{NMT}(p^{\prime}|p_t)$ ($p_{NMT}$ is the generation probability output by $f_{NMT}$) is selected as the paraphrased prompt template. The manual and paraphrased prompt templates are presented in Tab.~\ref{tab:prompt}. } 

\subsection{Contextualized Soft Token Model}
\label{sec:con_soft}

Generally, our contextualized soft token model belongs to the continuous prompt methods~\cite{liu2023pre}. Recall that one of the difficulties in the GEM problem lies in the noisy data, which may interfere with the PLM to produce poor results. We need to use suitable prompts to extract critical information from the entities or guide the LM to distinguish which part of the information is crucial for matching the two entities. In most previous continuous soft token approaches~\cite{wang2022promptem, liu2023gpt}, the soft tokens are tuned for a task instead of each data instance, which is insufficient for handling the aforementioned two features.

\bl{Therefore, we propose the contextualized soft token model that extracts information from each entity, which is provided to the LM to relieve such an issue. Intuitively, comparing different critical aspects of entities can help boost performance. For example, the relatively important information in a product entity includes the title, product identifier, and some key device parameters. We expect our soft token to learn to extract these features, which can be directly presented to the LM.}

\bl{In detail, we extend the previous serializing method in Sec.~\ref{sec:natural_prompt} with the soft tokens as follows:}
\begin{equation}
    \bl{\text{serialize\_soft}(e)=\text{serialize}(e)\ \text{the keyword is } [S_1]\dots [S_K],}
\end{equation}
\bl{where $[S_1]\dots[S_K]$ represents the soft tokens, and $K$ is the hyper-parameter that controls the number of tokens added. Each of the $K$ different tokens is expected to represent a different aspect extracted from the entities. We will present how to compute the embeddings of these $K$ tokens following.}

We first encode the input entity using the standard self-attention transformer module~\cite{vaswani2017attention}. Suppose that for an entity $e$, $\{t_1,...,t_l\}$ are the tokens output by the tokenizer of the text sequence $serialize(e)$. We follow the standard transformer processing pipeline (token embedding plus position embedding input) and use $N$ self-attention transformer layers to encode $\{t_1,...,t_l\}$ to $E_{1:l}$, where $E_{1:l}$ represents the output of embeddings after $N$ transformer layers. 

\bl{To extract $K$ aspects from a given entity, we aim to fetch the most semantically relevant tokens for the $K$ different aspects from the entity and aggregate them. So the aggregated embeddings are the semantic centers we seek to learn. Specifically, we use an attention module to extract different aspects of information from the given entity. We propose $A_{1:K} \in \mathcal{R}^{K\times d_q}$ as $K$ learnable aspect parameters, where $d_q=768$ is this parameters' length. After transforming $E_{1:l}$ to the key matrix $M_K \in \mathcal{R}^{l\times d_q}$ and the value matrix $M_V \in \mathcal{R}^{l\times d_v}$ ($d_v=768$ is the value matrix's length) using two linear layers, we conduct the scaled dot-product attention computation. The aggregated features, according to the attention matrix, are:}
\begin{equation}
    Emb_{[S]}=\text{softmax}(\frac{A_{1:K}M_K^T}{\sqrt{d_q}})M_V.
\end{equation}
After the attention layer, the soft token embedding undergoes post-processing, which includes linear transformations and normalization. The whole pipeline of our contextualized soft token model is presented in the lower-right part of Fig.~\ref{fig:enter-label}.

\subsection{Training Part of Prompt Tuning}
\label{sec:train}
During training, we choose the Prompt+LM tuning paradigm, tuning both the LM and prompt parameters, which fully activates the model capabilities~\cite{liu2023pre}. The LM part is tuned because the main LM module requires sufficient training to handle the GEM task~\cite{paganelli2022analyzing}. The prompt part is definitely tuned because a learnable soft token model is involved. Though the cost of Prompt+LM tuning is usually considered high, the cost can be neglected due to limited training samples, as we are dealing with the low-resource GEM task.

We follow the label word selection in PromptEM.  The conditional probability of an entity pair and its label is defined as the sum of the normalized scores in the corresponding label set with \{matched, similar, relevant\} as the word set for the matching label, and \{mismatched, different, irrelevant\}  for the non-matching label.

We propose an orthogonal regularization loss to guide the model in learning different soft embeddings, as we expect our learned parameters to represent diverse key aspects of the given entity. The orthogonal loss is computed as:
\begin{equation}
    \mathcal{L}_{ortho}=||Emb([S])Emb([S])^T-I||/d_s,
\end{equation}
where $S \in \mathcal{R}^{K\times d_s}$ and $I$ is the identity matrix. 

The orthogonal loss is combined with the normal cross entropy classification loss $\mathcal{L}_{CE}$ to be our prompt tuning loss:
\begin{equation}
    \mathcal{L}=\mathcal{L}_{CE}+\lambda\mathcal{L}_{ortho},
\end{equation}
in which $\lambda$ is a balance hyper-parameter. The choice of $\lambda$ is further discussed in Sec.~\ref{sec:exp}.
\begin{figure}
    \centering
    \includegraphics[width=0.48\textwidth]{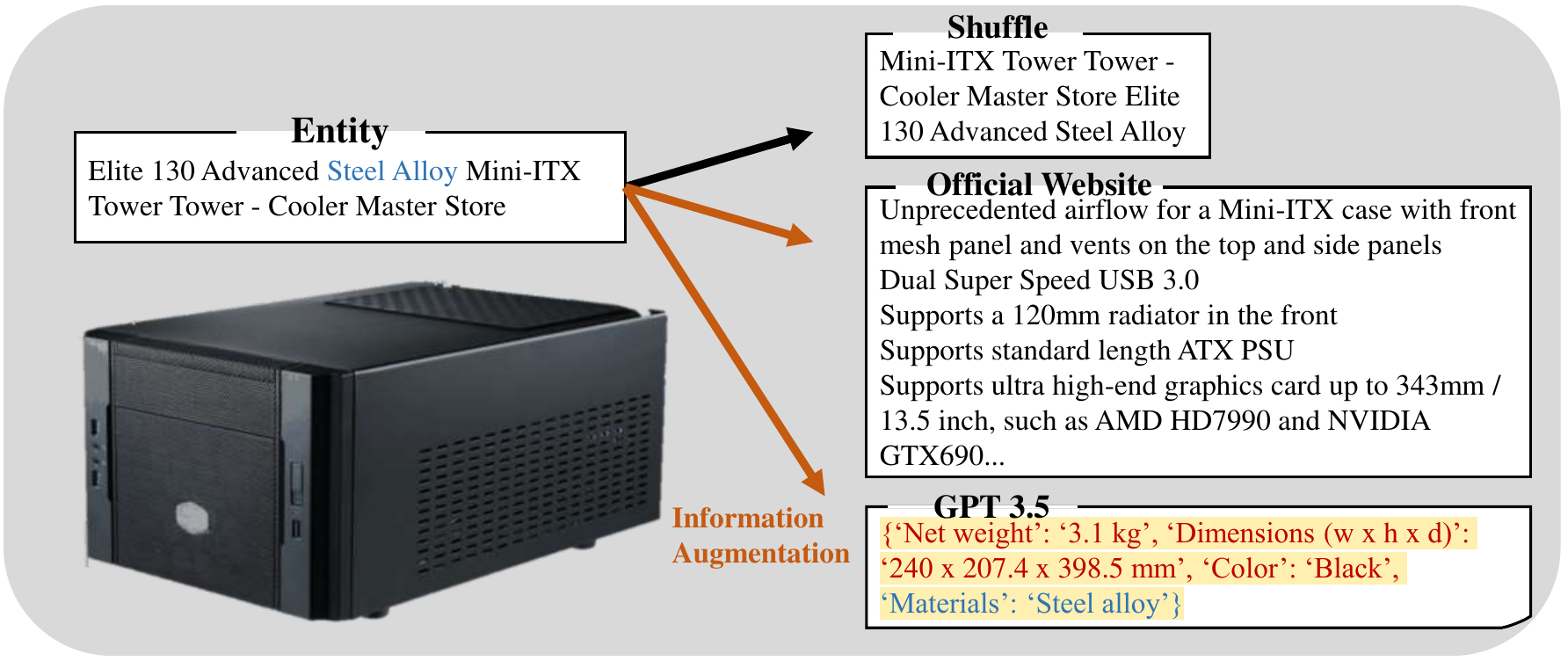}
    \caption{\bl{Example of different data augmentation operators. The shuffle operator from Ditto doesn't provide additional information, while the augmentation from GPT 3.5 can supplement new information (red part), and extract information in structured forms from the description (blue part).}}
    \label{fig:aug}
\end{figure}
\section{Information Augmentation for GEM}
\label{sec:info}

In this section, we first introduce the concept of `information augmentation', offering supplementary information, which can serve as a powerful data augmentation operator. Subsequently, we propose using LLMs to implement information augmentation. Lastly, we demonstrate the cost-effectiveness of this module and propose our uncertainty-based augmentation strategy to further reduce the LLM's token fee.

\bl{\subsection{Information Augmentation as a Powerful Data Augmentation Operator}}

\label{sec:i as d}
\bl{Many data augmentation operators have been proposed to help boost EM's performance and stability. The data augmentation methods can help models learn invariant transformations, guide models to neglect data noises, and serve as a regularization method to prevent the models from over-fitting~\cite{li2021data,feng2021survey}. For instance, Ditto proposes several data augmentation operators for EM (e.g., del, shuffle, swap, and the shuffle operator is presented in Fig.~\ref{fig:aug}) and interpolates the augmented data with the original data for training. Rotom~\cite{miao2021rotom} introduces a pre-trained (on rule-based augmented data) seq2seq model to perform personalized data augmentation. Sudowoodo~\cite{wang2023sudowoodo} and SupCon-EM~\cite{peeters2022supervised} perform data augmentation to provide training samples for self-contrastive learning. These data augmentation approaches provide rule-based invariant knowledge and other benefits. However, they typically bring no entity-aware information as no extra information source is introduced. }

\bl{Motivated by a series of retrieval-based LM approaches (e.g., KNN-prompt~\cite{shi2022knnprompt}, REPULG~\cite{shi2023replug}), which have shown that relevant information retrieved from proper data sources can boost the LM's few-shot and zero-shot capabilities dramatically~\cite{asai2023retrieval}, we propose the information augmentation operator. Information augmentation, which provides additional information for an entity, serves as a new data augmentation operator. It's a powerful data augmentation operator, as it introduces additional information.}

\bl{There can be various approaches to augmenting information, e.g., knowledge graphs or open internet data sources. For example, KaER~\cite{fang2023kaer} and KnowPrompt~\cite{chen2022knowprompt} retrieve information from knowledge graphs and integrate this information into the language model. However, knowledge graphs may suffer from issues of data currency and relevance, as rapidly changing information may not be constantly updated. Besides, locating specific entities within a knowledge graph is challenging. For example, the digital product entity in Fig.~\ref{fig:aug} is not likely to appear in a knowledge graph. Retrieving from open internet sources also suffers from high costs (API fees or manual effort) and legal risks~\cite{yu2020summary}.}

\bl{After PLMs are proposed, multiple studies have shown the capability of LMs as knowledge bases~\cite{petroni2019language,alkhamissi2022review}. Recently, the introduction of LLMs~\cite{openai2023gpt4} has elevated the performance of LM's fact probing capabilities to unprecedented levels. Though the augmentation of entities, such as songs, academic papers, and products, belongs to the long-tail distribution, which LMs are not familiar with, the large training corpus for LLMs relieves such an issue. For long-tail distributed data that LLMs have not been trained on, LLMs can also make guesses based on common sense~\cite{openai2023gpt4}.  Additionally, LLMs can extract useful information from the entity descriptions in a well-structured and brief format, which is more friendly to our prompt tuning framework. }

\bl{\subsection{LLM-enhanced Information Augmentation}}
\label{sec:llmenhance}

\bl{As we have mentioned above, since LLMs can augment information in structured forms, we should first determine which attributes are needed for EM. Then, we should design a prompt used for LLM queries and consider how to integrate new information into prompt tuning.}

{\bl{\stitle{Attribute Selection.}} \bl{Considering the heterogeneity of different data sources, we have to determine the \{attribute\_list\} for each dataset uniquely, where \{attribute\_list\} is a list of attribute names we desire for a specific entity. We propose two different ways of determining the attribute keys of an entity.}

\bl{I. Augmentation at the granularity of the data source. For a highly homogeneous dataset, we can manually design the attributes that are essential for the alignment of two entities within it. Advice from ChatGPT can also be taken into account. We can query ChatGPT for the attributes needed in the WDC dataset:}

\begin{tcolorbox}[colframe=black, colback=white, boxrule=1pt, sharp corners]
\noindent \textit{\bl{ List the features important for entity matching in a digital product dataset.}}
\end{tcolorbox}
\noindent \bl{Here, the \{attribute\_list\} for the WDC dataset can be \{ capacity, color, frequency, keywords, language, model number, product identifier, release year, resolution, size, speed, weight \}. The important attributes in the list can be either extracted or recalled by LLMs.} 

\bl{II. Augmentation at the granularity of the data instance. For a heterogeneous dataset, such as an e-commerce dataset with both digital products and daily essentials, using a uniform attribute query list is not appropriate. Therefore, the desired augment attributes should be determined by the entities to be matched for a given entity. As mentioned in Sec.~\ref{sec:intro}, a blocking pipeline is usually required before establishing a matching system for entity pairs. Assume that after the blocking system, the entities to be matched with entity $e$ are $e_1,...,e_n$. For each $e_i$, $KEY_i$ is the set of attribute keys $e_i$ has. (For nested semi-structured entities, attribute keys at every level should be considered.). The query attribute set for entity $e$ can be defined as:  $attribute\_list_e=\bigcup_{i=1}^{n} KEY_i$, so that the key attributes of other entities, that should be compared with entity $e$, can be collected. }

\bl{\stitle{Prompt Design.} We use the following prompt to augment structured information from the LLM:}

\begin{tcolorbox}[colframe=black, colback=white, boxrule=1pt, sharp corners]
 \textit{\bl{
[\{"role": "system", "content": "You are a helpful assistant. Answer in plain json format only"\},\\
\{"role": "user","content": "Please provide some information about the following entity. The entity is \{entity\_info\}. Please output the \{attribute\_list\} of the entity.")\}]
}}
\end{tcolorbox}
 \noindent \bl{where \{entity\_info\} is the entity data presented in text (semi-structured data transformed in JSON format). }

\stitle{Integration with Prompt Tuning.} As shown in Fig.~\ref{fig:enter-label}, the augmented information in JSON format is serialized as described in Sec.~\ref{sec:natural_prompt}, and the serialized sequence is concatenated with the original serialized sequence for the following computation. It is expected that specific desired attributes in \{attribute\_list\}do not exist for an entity, so we use regular expressions to match meaningless augmented information generated by LLMs. The value terms are then substituted by <pad> in the serialization.

\stitle{Cost for Augmenting using LLMs.} The API fee of original/fine-tuned GPT3.5-turbo inference is \$0.002, 0.016/1k tokens, respectively, in 10/2023~\footnote{https://openai.com/pricing}, which is still expensive if we aim to match a large number of entities. So it's necessary to consider the cost of using LLMs.

Suppose that we are performing GEM between two sources of $N$ entities, and each entity has $N_k$ pairs of key and value, which consist of $L$ tokens on average. After the blocking system, each entity from one source has an average of $B$ counterparts to compare with. If we perform information augmentation on the $2*N$ entities, we require an input of all the $2N$ entities, which requires $2*N(N_k*L)$ tokens. We further receive an output of the augmented attributes. If each entity has an $\{attribute\_list\}$ with an average length $O_k$, we receive a total output of $2*N(O_k*L)$ tokens.

Since directly using LLMs is a way to bridge the semantic gap, we believe it's necessary to compare it with our information augmentation strategy. There are generally two ways to use these foundation models~\cite{narayan2022can,peeters2023using}. One is inference in a zero/few-shot (in-context learning) fashion; the other is traditional fine-tuning. Augmenting information from LLMs can help boost our matching model's performance to be comparable to LLMs' performance, which will be presented in the Sec.~\ref{sec:exp}. So, considering that the performance of the models is similar, the cost issue becomes more critical. 

Augmenting information using LLMs generally costs less than directly using LLMs. A straightforward explanation for the cost-saving of LLM tokens when adding supplementary information is that during supplementation, all entities are only traversed once, while directly using LLM for comparison requires multiple traversals of all entities because a single entity may have multiple corresponding entities to be judged against. Here, directly using LLMs means inference from a pre-provided/fine-tuned LLM model, which is the major cost of using LLMs, given that we are solving low-resource GEM (small training samples). Besides, inference is indeed a recurring expense. Direct inference from an LLM approximately requires an input of all the entity pairs, which consists of $N*B*(2*(N_k*L))$ tokens. In comparison, we subtract the two costs, and the difference is therefore:
\begin{equation*}
    d=2*N*L(B*N_k-N_k-O_k).
\end{equation*}
In practice, $B*N_k$ is usually higher than $O_k$, which makes information augmentation more cost-effective. Suppose that the hyper-parameter $B$ of the blocking system is $5$, and the basic attribute number and augmented attribute number are the same, augmenting information using LLMs can save $60\%$ of the tokens used by direct inference.

 \subsection{Uncertainty-based Cost-effective Information Augmentation Strategy}
\label{sec:eco}
In this subsection, we propose an uncertainty-based information augmentation strategy that further reduces the API fee. It is a common observation that for some entities, entities from another source are easy to distinguish from, while for other entities, the matching is quite confusing. This outcome may result from the dataset's characteristics, or it could be caused by the information loss of the data instance. Leveraging this feature, we propose neglecting the entities that are certain in matching without the information augmentation procedure to prevent unnecessary API token costs.

During training, we augment all training instances to ensure that the matching model has the best performance, as the fees can be neglected due to the low-resource settings. During inference, we augment some of the test instances according to the uncertainty score. For the entity not augmented, we pad all the attribute values we query to <pad>. We select instances with similar classification scores, as they are confusing to the current model. Following the notations in Sec.~\ref{sec:llmenhance}, we define the uncertainty score of an entity $e$:
\begin{equation}
    Score(e)=max_{e_i\in \{e_1,...,e_n\}} H(P(y|(e,e_i))),
\end{equation}
where $\{e_1,...,e_n\}$ are the ones that the blocking module has selected as potential matches for entity $e$, and $H$ denotes the negative entropy calculation or the uncertainty metric $max(P(y=0|(e,e_i)),P(y=1|(e,e_i))$. The higher the score an entity has, the more confident the current model is about this entity. Then, only the entities with scores lower than $\tau$ are augmented using LLMs, in which $\tau$ represents an uncertainty threshold.

Though augmenting only part of the test data may damage the consistent data distribution between the training set and test set, we find through experiment that the strategy has little impact on model performance and saves many API tokens. This is because we convert all entities to natural language, especially using reverse translation techniques, ensuring that the data distribution between augmented and original data instances remains similar.
\begin{table*}[]
\caption{\bl{Dataset Statistics: The schemas with underlines are relatively long textual attributes, the schemas in braces \{\} are attributes in iterative dict format, and the schemas in square brackets [] are attributes in list format. The left and right schemas are separated by a vertical bar |. The left and right sources share the same schema if | is missing. }}
\begin{tabular}{l|lll|l|l}
\hline
Dataset              & \#Left & \#Right & \#Labels & Domain     & Data Schema                                                                                                                                                                                                   \\ \hline
{REL-TEXT}    & 2,616       & 2,295        & 7,417    & Academic   & \bl{\underline{abstract} | title, authors, venue, year }                                                                                    \\ \cline{6-6} 
{SEMI-HETER}  & 22,133      & 23,264       & 1,240    & Book       & \begin{tabular}[c]{@{}l@{}} \bl{\{publication info\} |  title,  authors,  \{publication info\}}\end{tabular}         \\ \cline{6-6} 
{GEO-HETER}   & 2,469       & 2,788        & 2,500    & Location   & \begin{tabular}[c]{@{}l@{}} \bl{name, latitude, longtitude, address, postal code}\\ \bl{| name, position, address, postal code}\end{tabular}                                                                             \\ \cline{6-6} 
{SEMI-HOMO}   & 2,616       & 64,263       & 17,223   & Academic   & \bl{title, [authors], venue, year}                                                                                                                                                                               \\ \cline{6-6} 
{SEMI-TEXT-W(S-T-W)} & 9,234       & 9,234        & 5,540    & Product    & \begin{tabular}[c]{@{}l@{}} \bl{category,  identifiers,  brand, \{key value pairs\}, price | \underline{info}}\end{tabular}                                                                               \\ \cline{6-6} 
{SEMI-TEXT-C(S-T-C)} & 20,897      & 20,897       & 12,538   & Product    & \begin{tabular}[c]{@{}l@{}}\bl{category,  identifiers,  brand,  \{key value pairs\}, price | \underline{info}}\end{tabular}                                                                               \\ \cline{6-6} 
{SEMI-REL}    & 29,180      & 32,823       & 1,309    & Movie      & \begin{tabular}[c]{@{}l@{}}\bl{title, director, [actors], year, ratings, \underline{information} | \{movie info\}}\end{tabular} \\

\cline{6-6} 
{\bl{REL-HETER}}   &  \bl{ 533}          & \bl{331}             &      \bl{567}    & \bl{Restaurant} & \begin{tabular}[c]{@{}l@{}}\bl{title, address, phone, category  | address, city, phone, type, class}\end{tabular}    
\\  
 
 \cline{6-6} {WDC}         & 1,215       & 1,220        & 2,500    & Product    &\bl{ brand, title, \underline{description}, price, price currency}                                                                                                                                                                                                                                                 \\  \hline
 \cline{6-6}
GOOGLE-AMAZON        & 1,363       & 3,226        & 9,167    & Product    & \bl{title, manufacturer, price  }                                                                                                                                                                              \\ \cline{6-6} 
\bl{WALMART-AMAZON}       & \bl{1,688}           &  \bl{5,247}             &   \bl{6144}       & \bl{Product}    & \bl{category, brand, model number, price}                                                                                                                                                                      \\ \cline{6-6} 
\bl{ITUNES-AMAZON }       &   \bl{262 }         &  \bl{436}            &   \bl{321}       & \bl{Music}      & \begin{tabular}[c]{@{}l@{}} \bl{song name, artist, album, genre, price, copyright,time, released date}\end{tabular}                                                                                              \\ \hline
\end{tabular}

\label{tab:dataset}
\end{table*}

\section{Experiment}
\label{sec:exp}
In this section, we conduct extensive experiments on real-world datasets to evaluate our proposed prompt tuning method and information augmentation module. Our objective is to address the following research inquiries through our experiments:

\begin{itemize}   
 \item I1: How does our prompt tuning method perform compared with other state-of-the-art EM methods in the low-resource GEM task? How does each module affect the model's performance?
\item I2: From which perspective does the contextualized soft token benefit GEM?
\item I3: How does our information augmentation module perform effectively and economically compared with directly using LLMs? How does this module help GEM?
\end{itemize}

 \subsection{Experimental Setup}

 \stitle{\bl{Datasets.}} \bl{We use a total of twelve real-world datasets to conduct our experiment. The dataset statistics are presented in Tab.~\ref{tab:dataset}. The first nine datasets are GEM tasks with complex and different data schemas, and the rest are three EM tasks with identical data schemas. For the GEM datasets, we use seven real-world GEM benchmarks from Machamp~\cite{wang2021machamp} and GEO-HETER from PromptEM~\cite{wang2022promptem}. Further, we include the WDC-product dataset~\cite{peeters2023wdc} newly collected in 12/2022, which makes it more likely to escape the LLM's training data. We use the hardest test cases of WDC, which is the test dataset with `$80\%$ corner cases' (hard positives and negatives) and all unseen cases (entities not appearing in training cases). For the traditional EM dataset, we select three datasets from the ER-Magellan dataset~\cite{kopcke2010evaluation}, GOOGLE-AMAZON, WALMART-AMAZON, and ITUNES-AMAZON datasets to show the boost in our performance on the traditional EM tasks.}

\stitle{Baselines.} We compare APrompt4EM with several representative EM methods, including state-of-the-art methods in low-resource GEM settings. The baselines compared are listed as follows:

\begin{itemize}
 \item \stitle{RoBERTa.} RoBERTa represents fine-tuning a RoBERTa~\cite{liu2019roberta} PLM to perform EM.
 \item \stitle{\bl{SentenceBert in EMNLP' 19~\cite{reimers2019sentence}.}} \bl{SentenceBert learns sentence embeddings using siamese pre-trained LMs to perform EM.}
 \item \stitle{Ditto in VLDB' 21~\cite{li2020deep}.} Ditto is a fine-tuning paradigm EM method that is improved by injecting domain knowledge, summarizing, and information augmentation.
 \item \stitle{Sudowoodo in ICDE' 23~\cite{wang2023sudowoodo}.} Sudowoodo is a self-contrastive learning framework that can learn representations for EM.
 \item \stitle{Rotom in SIGMOD' 21~\cite{miao2021rotom}.} Rotom is a meta-learned seq2seq data augmentation framework for data processing tasks including EM.
 \item \stitle{\bl{Machop in aiDM' 22~\cite{wang2022machop}.}} \bl{Machop is an EM framework with a structure aware pooling layer that compares each attribute pair.}
 \item \stitle{PromptEM in VLDB' 22~\cite{wang2022promptem}.} PromptEM is the state-of-the-art low-resource GEM solution, which is based on prompt tuning, enhanced by pseudo-labels and self-training. We mainly compare its prompt tuning module.
 \item \stitle{GPTx.} GPTx represents fine-tuning the OPENAI proposed LLM on the GEM task.
    
\end{itemize}

\stitle{Implementation Details.} All the backbone LMs of our method and the baselines are selected as RoBERTa. We implement our model based on the PromptEM source codes\footnote{https://github.com/ZJU-DAILY/PromptEM}, Pytorch~\cite{paszke2019pytorch}, the Transformers library~\cite{wolf2019huggingface}, and the OpenPrompt Library~\cite{ding2021openprompt}. 

\bl{The low-resource settings refer to using $10\%$ of the datasets' labeled data.} We enumerate through the hyper-parameters and report the best results. \bl{The soft token number $K$ is selected from $\{1,2,4,8\}$, the number of transformer encoder layers in the contextualized soft token is searched in $0-2$, and the hyperparameter $\lambda$ of the orthogonal loss is set to $1$.} The mode of position embeddings in the contextualized soft token is selected from \{COL, POS, w/o \}, in which COL means using different position embeddings for different columns~\cite{nassar2022tableformer}, PE means using original position embeddings, and w/o means not using position embeddings. Following the training settings of PromptEM, we train our model using the AdamW optimizer~\cite{loshchilov2017decoupled}. The number of training epochs is set to $30$, the learning rate is set to $2e-5$, and the batch size is set to $24$. The prompt template we used is selected from Tab.~\ref{tab:prompt}. The experiments are conducted on an Ubuntu 20.04.6 LTS Linux server with an Intel Xeon Silver 4210R CPU, $256$ GB of RAM, and an NVIDIA RTX 3090 graphics card. The RoBERTa, SentenceBert, Ditto, Sudowoodo, and Rotom methods are implemented through their open-source codes\footnote{https://github.com/megagonlabs}, and the hyper-parameters are set to the recommended ones. We implemented our own Machop baseline according to the model structures and hyper-parameters described in the original paper. The fine-tuning of GPTx models is implemented through the OPENAI official API using the default hyper-parameters.

\begin{table*}[ht]
% Please add the following required packages to your document preamble:
% \usepackage{graphicx}
\centering
\caption{Main results of comparing methods in the low-resource GEM setting.}
\resizebox{\textwidth}{!}{%
\begin{tabular}   {l|p{0.3cm}p{0.3cm}p{0.45cm}|p{0.3cm}p{0.3cm}p{0.45cm}|p{0.3cm}p{0.3cm}p{0.45cm}|p{0.3cm}p{0.3cm}p{0.45cm}|p{0.3cm}p{0.3cm}p{0.45cm}|p{0.3cm}p{0.3cm}p{0.45cm}|p{0.3cm}p{0.3cm}p{0.45cm}|p{0.3cm}p{0.3cm}p{0.45cm}|p{0.3cm}p{0.3cm}p{0.45cm}|p{0.3cm}p{0.3cm}p{0.45cm}|p{0.3cm}p{0.3cm}p{0.45cm}|p{0.3cm}p{0.3cm}p{0.45cm}}\hline
              & \multicolumn{3}{c|}{REL-TEXT}  & \multicolumn{3}{c|}{SEMI-HETER} & \multicolumn{3}{c|}{GEO-HETER} & \multicolumn{3}{c|}{SEMI-HOMO} & \multicolumn{3}{c|}{S-T-W} & \multicolumn{3}{c|}{S-T-C} & \multicolumn{3}{c|}{SEMI-REL}  & \multicolumn{3}{c|}{\bl{REL-HETER}} & \multicolumn{3}{c|}{WDC}  & \multicolumn{3}{c|}{G-A} & \multicolumn{3}{c|}{\bl{W-A}}& \multicolumn{3}{c}{\bl{I-A}}       \\ \hline
Methods       & P     & R     & F              & P      & R     & F              & P     & R     & F              & P     & R     & F              & P      & R      & F              & P      & R      & F              & P     & R     & F              & P      & R      & F              & P     & R     & F    & P     & R     & F   & P     & R     & F   & P     & R     & F             \\ \hline
RoBERTa & 19.8 & 35.1 & 25.3 & 25.3 & 40.9 & 31.3 & 83.9 & 83.3 & 83.6 & 94.5 & 93.0 & 93.7 & 30.0 & 1.4 & 2.7 & 76.6 & 71.9 & 74.2 & 43.3 & 95.1 & 59.5 & \bl{100} & \bl{95.5} & \bl{97.7} & 20.9 & 12.6 & 15.7 & 59.0 & 59.0 & 59.0 & \bl{56.2} & \bl{65.8} & \bl{60.6} & \bl{75.9} & \bl{81.5} & \bl{78.6}   \\
SentenceBert & 44.0 & 49.5 & 46.8 & 70.2 & 75.5 & 72.7 & 64.0 & 81.5 & 71.7 & 91.8 & 90.4 & 91.1 & 24.5 & 12.8 & 16.8 & 69.7 & 54.4 & 61.1 & 76.8 & 86.9 & 81.5 & \bl{100} & \bl{90.0} & \bl{94.7} & 25.7 & 55.0 & 35.1 & 37.5 & 64.1 & 47.3 & \bl{26.9} & \bl{47.2} & \bl{34.3} & \bl{76.7} & \bl{85.2} & \bl{80.7}
     \\
Ditto & 37.6 & 51.4 & 43.4 & 92.6 & 62.9 & 74.9 & 78.3 & 86.0 & 82.0 & 91.9 & 95.0 & 93.4 & 29.5 & 31.3 & 30.3 & 56.8 & 47.1 & 51.5 & 86.8 & 97.3 & 91.8 & \bl{100} & \bl{86.4} & \bl{92.7} & 11.7 & 87.8 & 20.7 & 50.3 & 73.1 & 59.6 & \bl{58.0} & \bl{54.4} & \bl{56.1} & \bl{71.4} & \bl{74.1} & \bl{72.7}
\\
Sudowoodo & 38.0 & 55.9 & 45.3 & 68.0 & 63.2 & 65.5 & 84.2 & 89.2 & 86.7 & 92.6 & 92.5 & 92.6 & 22.0 & 31.8 & 26.0 & 67.4 & 67.0 & 67.2 & 83.3 & 95.6 & 89.1 & \bl{96.0} & \bl{97.2} & \bl{96.6} & 47.2 & 60.2 & 52.9 & 50.4 & 58.5 & 54.2 & \bl{65.3} & \bl{65.3} & \bl{65.3} & \bl{91.2} & \bl{93.6} & \bl{92.3}
   \\
Rotom & 61.5 & 50.3 & 55.4 & 70.2 & 45.9 & 55.5 & 72.2 & 56.5 & 63.4 & 86.1 & 91.7 & 88.8 & 21.9 & 20.9 & 21.4 & 86.1 & 59.8 & 70.5 & 74.0 & 93.4 & 82.6 & \bl{98.6} & \bl{100} & \bl{99.3} & 17.8 & 14.0 & 15.7 & 70.2 & 45.9 & 55.5 & \bl{71.7} & \bl{62.3} & \bl{66.7} & \bl{92.1} & \bl{89.7} & \bl{90.9}
\\
\bl{Machop} & \bl{18.0} & \bl{99.2} & \bl{30.5} & \bl{73.6} & \bl{80.5} & \bl{76.9} & \bl{75.6} & \bl{88.1} & \bl{83.1} & \bl{92.7} & \bl{93.2} & \bl{93.0} & \bl{11.6} & \bl{99.1} & \bl{20.7} & \bl{18.6} & \bl{48.2} & \bl{26.8} & \bl{90.4} & \bl{97.8}  & \bl{94.0} & \bl{100} & \bl{86.4} & \bl{92.7} & \bl{22.2} & \bl{25.4} & \bl{23.7} & \bl{55.6} & \bl{74.8} & \bl{63.8} & \bl{68.2} & \bl{68.9} & \bl{68.6} & \bl{84.6} & \bl{40.7} & \bl{55.0}
    \\
PromptEM & 54.9 & 62.8 & 58.6 & 94.0 & 39.6 & 55.8 & 82.9 & 86.5 & 84.7 & 94.2 & 95.0 & 94.6 & 32.9 & 22.8 & 26.9 & 86.2 & 73.4 & \textbf{79.3} & 94.5 & 94.5 & 94.5 & \bl{100} & \bl{100} & \bl{\textbf{100}} & 19.9 & 44.4 & 27.5 & 65.6 & 60.3 & 62.8 & \bl{81.3} & \bl{70.0} & \bl{75.2} & \bl{100} & \bl{92.6} & \bl{96.2}
      \\ \hline
APrompt4EM & 66.5 & 63.1 & \textbf{64.7} & 81.6 & 78.0 & \textbf{79.7} & 83.7 & 90.2 & \textbf{86.8} & 94.2 & 96.5 & \textbf{95.3} & 76.0 & 52.6 & \textbf{62.2} & 84.9 & 73.1 & 78.6 & 92.8 & 98.9 & \textbf{95.8} & \bl{100} & \bl{100} & \bl{\textbf{100}} & 53.6 & 56.6 & \textbf{55.1} & 63.6 & 70.9 & \textbf{67.1} & \bl{80.6} &\bl{79.8} & \bl{\textbf{80.2}} & \bl{100} & \bl{100} & \bl{\textbf{100}}
\\
\bl{w d-hyp} & \bl{59.9} & \bl{65.3} & \bl{62.5} & \bl{81.1} & \bl{73.8} & \bl{77.3} & \bl{86.7} & \bl{85.1} & \bl{85.9} & \bl{95.1} & \bl{94.6} & \bl{94.9} & \bl{68.1} & \bl{46.5} & \bl{55.2} & \bl{83.6} & \bl{71.2} & \bl{76.9} & \bl{93.1} & \bl{95.6} & \bl{94.3} & \bl{100} & \bl{95.5} & \bl{97.7} & \bl{46.6} & \bl{50.2} & \bl{48.3} & \bl{55.1} & \bl{71.8} & \bl{62.3} & \bl{83.1} & \bl{76.2} & \bl{79.5} & \bl{100} & \bl{100} & \bl{\textbf{100}}
  \\
w/o reg. & 65.2 & 63.3 & 64.2 & 79.4 & 67.9 & 73.2 & 87.4 & 84.4 & 85.9 & 94.8 & 94.3 & 94.6 & 76.0 & 52.6 & \textbf{62.2} & 87.6 & 68.4 & 76.8 & 92.8 & 97.8 & 95.2 & \bl{100} & \bl{100} & \bl{\textbf{100}} & 49.1 & 58.4 & 53.4 & 58.3 & 64.5 & 61.3 & \bl{78.7} & \bl{68.9} & \bl{73.5} & \bl{100} & \bl{100} & \bl{\textbf{100}}
    \\
w/o soft & 60.6 & 68.9 & 64.5 & 80.5 & 59.8 & 68.6 & 84.3 & 82.4 & 83.3 & 92.5 & 96.0 & 94.2 & 20.5 & 36.5 & 26.2 & 77.5 & 73.2 & 75.3 & 92.1 & 95.1 & 93.6 & \bl{100} & \bl{100} & \bl{\textbf{100}} & 36.6 & 50.6 & 42.5 & 52.7 & 63.3 & 57.5 & \bl{80.1} & \bl{73.1} & \bl{76.4} & \bl{100} & \bl{100} & \bl{\textbf{100}}
 \\
w/o NL & 59.3 & 65.3 & 62.2 & 81.9 & 71.3 & 76.3 & 81.2 & 86.7 & 83.9 & 93.1 & 96.0 & 94.5 & 59.5 & 31.3 & 41.0 & 80.3 & 68.9 & 74.2 & 91.2 & 96.7 & 93.9 & \bl{95.7} & \bl{100} & \bl{97.8} & 43.2 & 62.0 & 50.9 & 51.0 & 75.2 & 60.7 & \bl{73.5} & \bl{80.3} & \bl{76.7} & \bl{93.1} & \bl{100} & \bl{96.4}
        \\ \hline
\end{tabular}%
}

\label{tab:main_result}
\end{table*}

\subsection{Results for our Prompt Tuning Model. (I1)}

\stitle{Main Results.} We compare the basic APrompt4EM without information augmentation with other methods in this subsection. The overall performance under the low-resource GEM task of all involving models is presented in the upper part of Tab.~\ref{tab:main_result}. On all twelve datasets, our model achieves the best results in eleven of them, and we achieve comparable results on the last dataset, S-T-C. Compared with the best results of baselines, our prompt tuning approach achieves an average of $5.24\%$ performance boost. Three main conclusions can be drawn according to the experiment results. \stitle{I.} Models of the fine-tuning paradigm (RoBERTa, Ditto, Machop) and contrastive pre-train paradigm (Sudowoodo) mostly perform worse than models of the prompt tuning paradigm (PromptEM, APrompt4EM), which verifies our choice of selecting prompt tuning as our base method. As stated in PromptEM~\cite{wang2022promptem}, the prompt tuning paradigm can transfer and adapt information from LMs better than directly fine-tuning the original LM. \stitle{II.} Compared with the prompt tuning method PromptEM, APrompt4EM performs better, as the natural language prompt is closer to the LM's text distribution, and the contextualized soft tokens extract the key information. Especially on datasets (SEMI-HETER, WDC, and S-T-W), which are comparably noisy and full of redundant information, our model can achieve a significant \textbf{$20\%+$} F1 improvement compared with PromptEM. \stitle{III.} APrompt4EM outperforms Rotom consistently on the low-resource GEM task. Compared with the Seq2seq data augmentation module in Rotom, the contextualized soft token can compress and extract information better.

\stitle{Ablation Studies.}
The ablation studies of the key components of the prompt tuning module are presented in the lower part of Tab.~\ref{tab:main_result}. \bl{w d-hyper represents our model with a default $N = 0, K = 4$ hyperparameters.} w/o reg. represents our method of removing the orthogonal loss, w/o soft represents our method of removing the contextualized soft token model, and w/o NL represents removing our natural language serialize module and using the structured serialize method proposed by Ditto.

\bl{As we can see, the results of w d-hyper slightly decline compared with the results of the optimized hyperparameters, but APrompt4EM with default hyperparameters still achieves a performance boost compared with the baselines.} For the other three ablations, we find out that all three key components are crucial to our final performance. The orthogonal loss could prevent all the soft tokens from being identical, which is helpful in all datasets. The contextualized soft token works mainly on the most redundant dataset, S-T-W, and the natural language module helps improve most on the S-T-W, as the structured data for one source differs severely from the textual data for another source, which harms the performance using the structured prompt template.
\begin{figure}[htp]
\centering
\begin{tabular}[t]{cc}
\subfigure[GOOGLE-AMAZON   ]{
      \psfig{figure=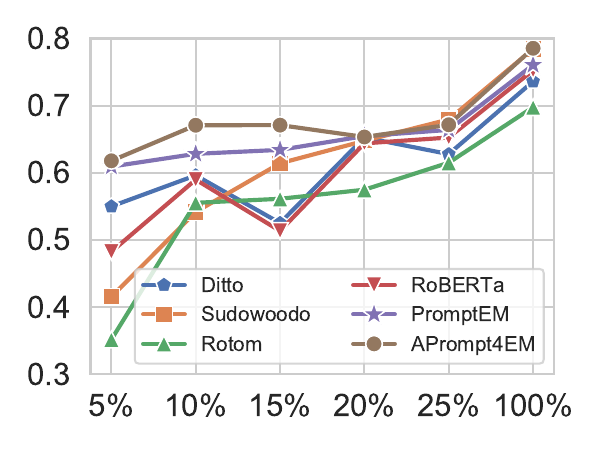,width=0.45\linewidth }
      \label{fig:dblp_k}
      }
      
\subfigure[WDC]{
      \psfig{figure=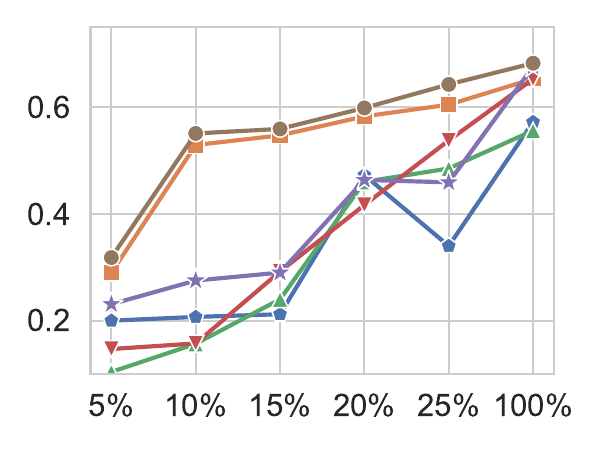,width=0.45\linewidth }
      \label{fig:twitter_k}
      }
\end{tabular}
%\vspace{-0.4cm}
\caption{\bl{F1 results with varying training ratio.}} 
\label{fig:trainratio}
%\vspace{-0.5cm}
\end{figure}

\stitle{\bl{Effectiveness in Different Low-Resource Settings.}} \bl{Our default low-resource settings use $10\%$ of the labeled training data for each dataset. We test the effectiveness of our model on the GOOGLE-AMAZON and WDC datasets from $5\%$ to $100\%$ of the training data. The results are shown in Fig.~\ref{fig:trainratio}. As we can see, our method shows promising results in general low-resource settings. The results of fine-tuning methods (RoBERTa, Ditto) show significant fluctuation when the training ratio is relatively small, which indicates that when the training set is tiny, the LM may easily be overfitting while fine-tuning, leading to unstable results. When the ratio of training data continues to rise, the gap between prompt tuning and fine-tuning methods gradually narrows, which is consistent with the hypothesis that prompt-tuning has more advantages in low-resource settings~\cite{lester2021power}. When $100\%$ of the labels are used, our method is also among the best-performing models, but the results are very close to the contrastive learning approach (Sudowoodo).}
\begin{figure}[htp]
\centering
\begin{tabular}[]{cc}

\subfigure[GOOGLE-AMAZON   ]{
      \psfig{figure=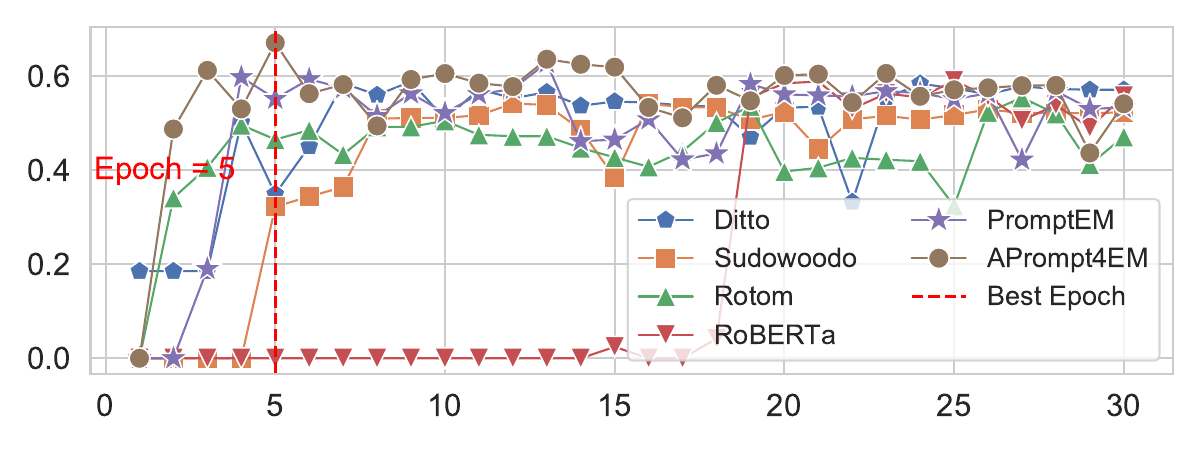,width=0.8\linewidth }
      \label{fig:dblp_k}
      }
      \\
\subfigure[WDC]{
      \psfig{figure=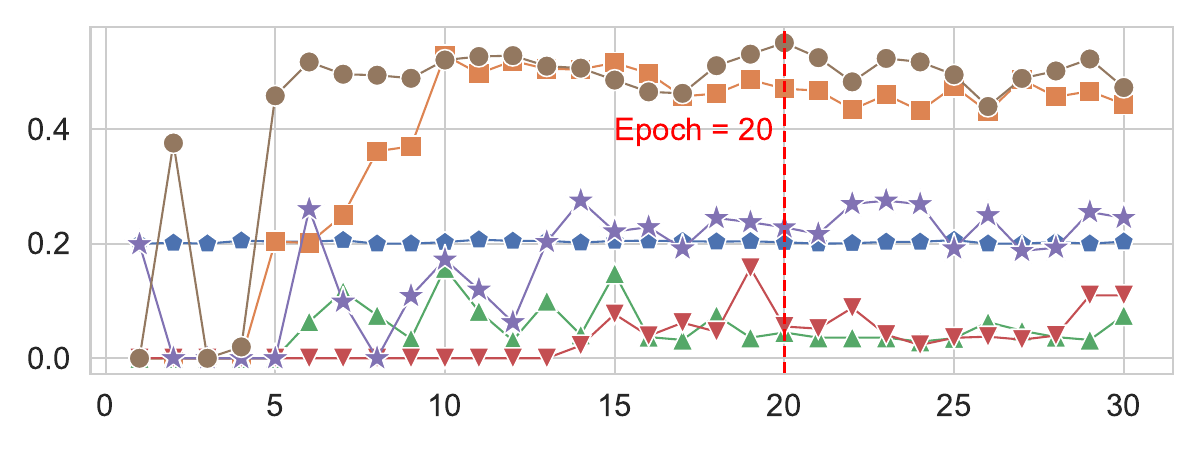,width=0.8\linewidth }
      \label{fig:twitter_k}
      }
      \\
\subfigure[SEMI-HOMO]{
      \psfig{figure=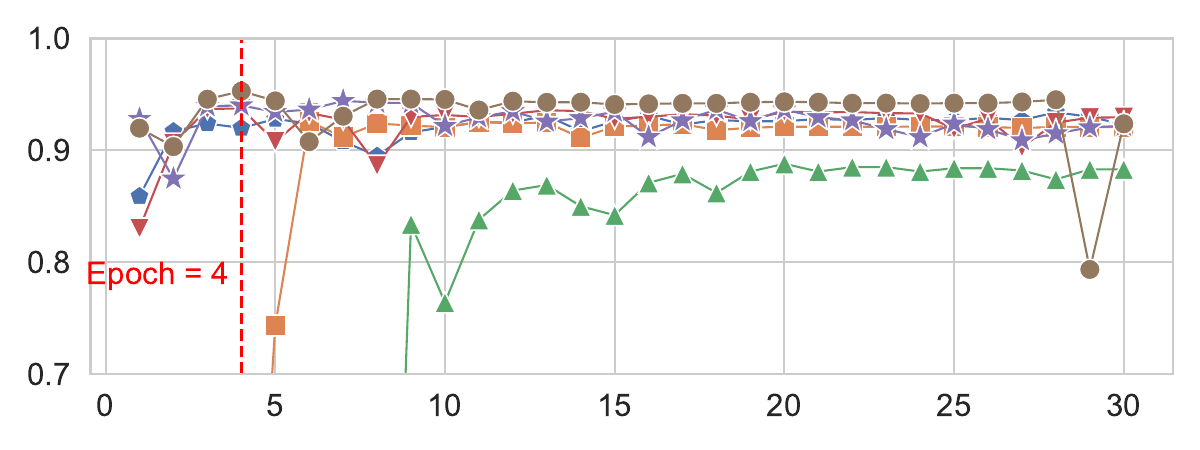,width=0.8\linewidth }
      \label{fig:twitter_k}
      }

\end{tabular}
%\vspace{-0.4cm}
\caption{F1 results with varying training epoch. } 
\label{fig:trainingepoch}
%\vspace{-0.5cm}
\end{figure} 

\stitle{Faster Convergence.} Through this experiment, we demonstrate that our natural language prompt module can help our model converge faster. We present the results of F1 on the test set for three datasets: GOOGLE-AMAZON, WDC, and SEMI-HOMO in Fig.~\ref{fig:trainingepoch}. As we can see, our method of using natural language prompts converges consistently faster than PromptEM and achieves the best results in a relatively early stage of the training process (Epoch $5$, $20$, and $4$ out of $30$ total epochs, respectively). More importantly, our natural language prompts can achieve a decent result in the first two epochs of training. For example, in GOOGLE-AMAZON, where both data sources are structured tables, the structured template makes the text sequence lose meaning when the model parameters are not tuned. Converting them to natural language ensures meaning for each entity without training, which can be captured by the LM in the very early stage of training.

\subsection{Further Experiments of the Contextualized Soft Token. (I2)}
\begin{figure*}[htp] 
\centering
\includegraphics[width=0.98\linewidth]{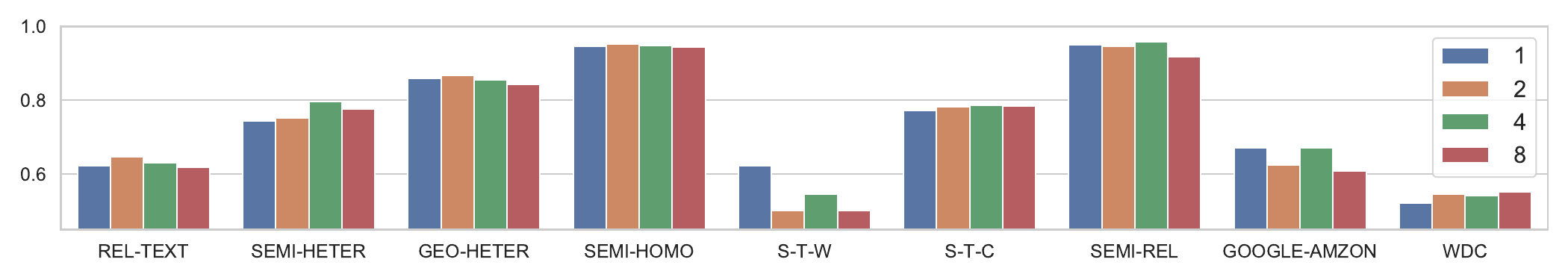}
\caption{F1 results with varying soft token number $K$.}
\label{fig:querysize}
\end{figure*}
\stitle{Effect of different hyper-parameter variants.} We present the performance of contextualized soft tokens with varying hyper-\\parameters to gain a deeper understanding of our key module. The F1 scores for different numbers of soft tokens $K$ across all datasets are depicted in Fig.~\ref{fig:querysize}. The results illustrate that for most datasets, employing multiple soft tokens/aspect queries is beneficial. This shows that, with the guidance of the orthogonal loss, multiple soft tokens might represent different aspects compressed from the entity, which is more helpful to our backbone LM. Moreover, the results of $8$ tokens are not always the best, which shows that the number of soft tokens follows `not more is better'. Selecting a proper soft token number can help the contextualized soft token extract information better.
\begin{figure}[htp]
\centering
\begin{tabular}[t]{cc}
\subfigure[GEO-HETER ]{
      \psfig{figure=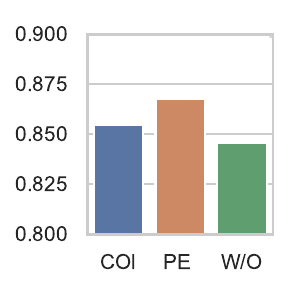,width=0.3\linewidth }
      \label{fig:dblp_k}
      }
\subfigure[SEMI-HETER]{
      \psfig{figure=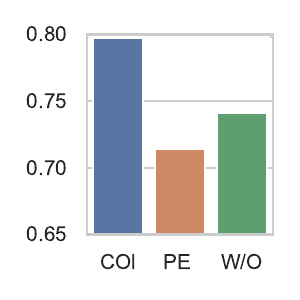,width=0.3\linewidth }
      \label{fig:twitter_k}
      }
\subfigure[SEMI-REL]{
      \psfig{figure=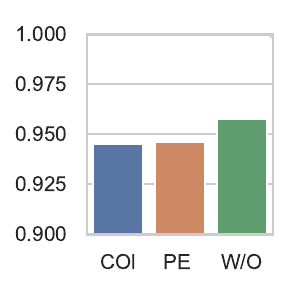,width=0.3\linewidth }
      \label{fig:twitter_k}
      }
\end{tabular}
%\vspace{-0.4cm}
\caption{F1 results with varying position embedding selection in the contextualized soft token module. } 
\label{fig:posemb}
%\vspace{-0.5cm}
\end{figure} 
We present the F1 scores of the GEO-HETER, SEMI-HETER, and SEMI-REL datasets under different position embedding settings in Fig.~\ref{fig:posemb}. The results vary for different datasets. For GEO-HETER, the location dataset and the location names, which are sensitive to the token orders, are better encoded using the original position embedding (PE). For SEMI-HETER having both structured data sources, the column position embedding (COL) is most suitable. As for SEMI-REL having sources with different data schemas, the difference may make no position embedding (w/o) better.
\begin{figure}[htp]
\centering
\begin{tabular}[t]{cc}
\subfigure[GOOGLE-AMAZON]{
      \psfig{figure=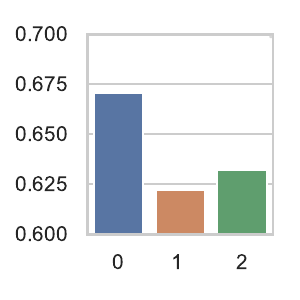,width=0.3\linewidth }
      \label{fig:dblp_k}
      }
\subfigure[SEMI-HETER]{
      \psfig{figure=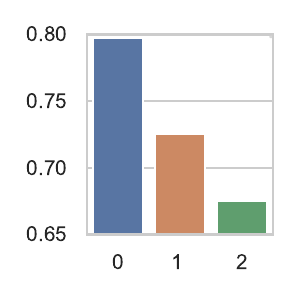,width=0.3\linewidth }
      \label{fig:twitter_k}
      }
\subfigure[S-T-C]{
      \psfig{figure=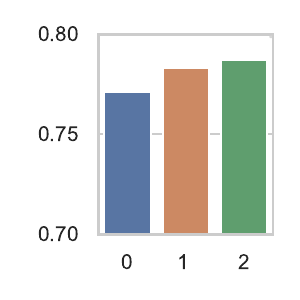,width=0.3\linewidth }
      \label{fig:twitter_k}
      }
\end{tabular}
%\vspace{-0.4cm}
\caption{F1 results with varying N layers.} 
\label{fig:encoderlayer}
%\vspace{-0.5cm}
\end{figure} 

Finally, we present the F1 results of modules with different transformers encoding layer number $N$ in Fig.~\ref{fig:encoderlayer}. As we can see, different datasets prefer different choices of $N$, as the functions of this module with different $N$ layers. We further show that in the following visualization part.

\begin{figure*}[]
\centering
\begin{tabular}[t]{cc}
 \hspace{-0.5cm}
\subfigure[WDC with 1 layer    (N=0)]{
      \psfig{figure=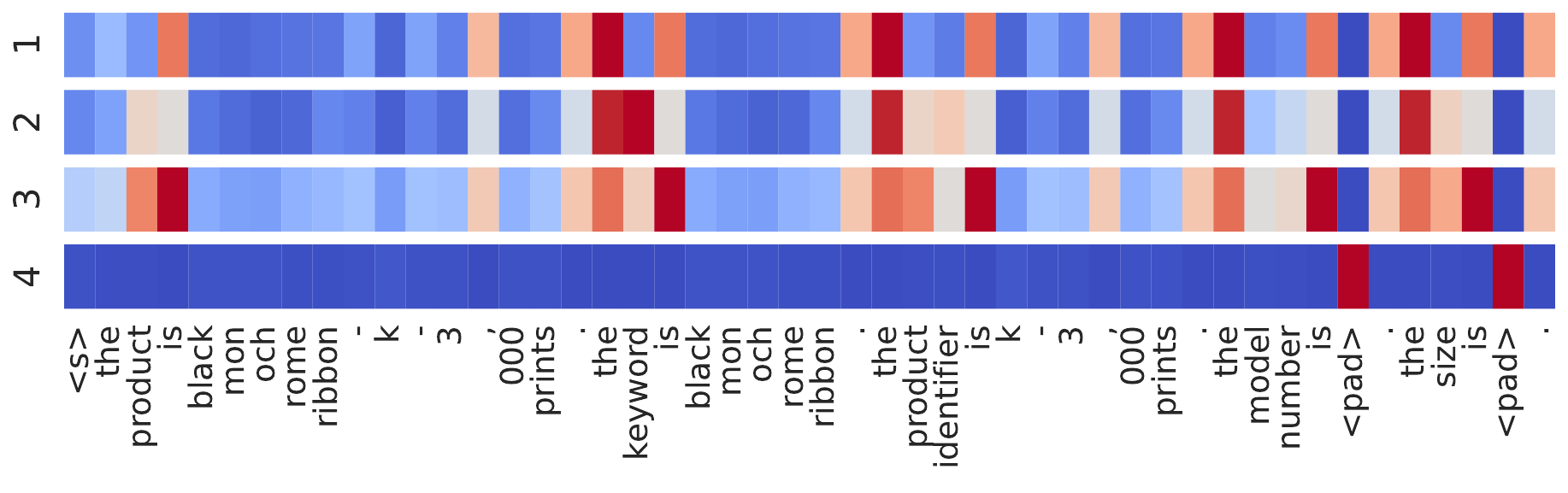,width=0.5\linewidth }
      \label{fig:dblp_k}
      }

\subfigure[WDC  with 2 layers (N=1)]{
      \psfig{figure=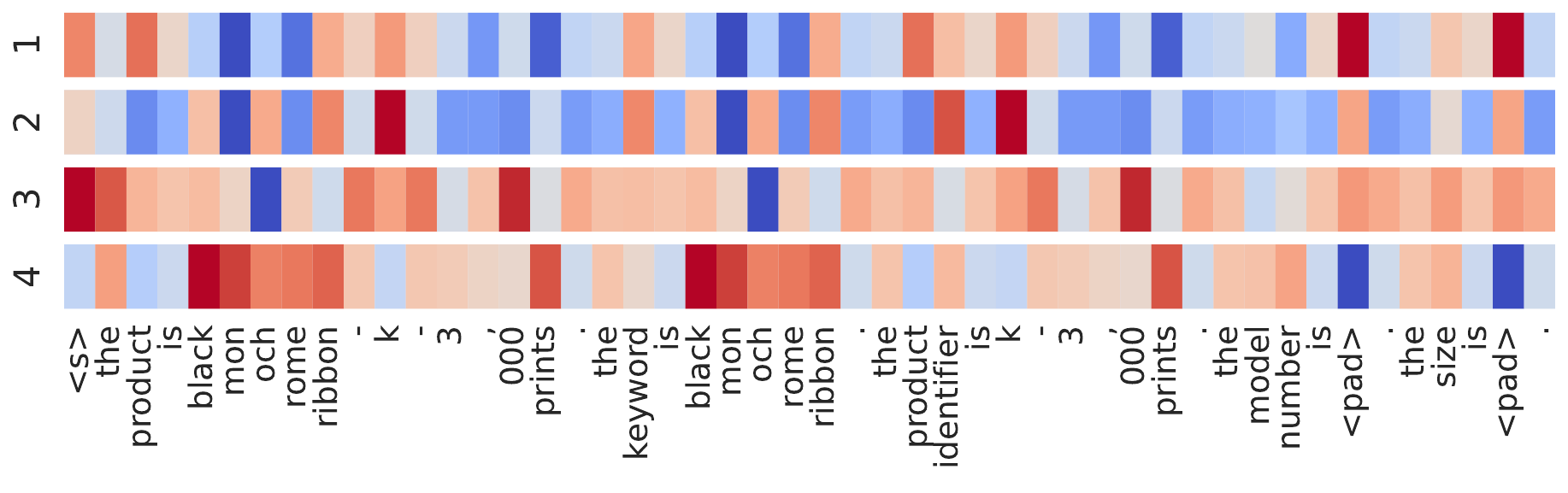,width=0.5\linewidth }
      \label{fig:twitter_k}
      }
\end{tabular}
%\vspace{-0.4cm}
\caption{\bl{Visualization of the attention between the aspect queries and the encoded tokens in the contextualized soft token module. We truncate the sentence due to page limitations. The redder the color, the greater the attention weight. }}
\label{fig:visual}
%\vspace{-0.5cm}
\end{figure*} 

\stitle{Visualization of the Contextualized Soft Token.} We visualize the attention score of $K=4$ aspect tokens with the encoded text tokens of the WDC dataset in Fig.~\ref{fig:visual}. The blue color corresponds to smaller attention values, whereas the red color indicates larger attention values. As we can see, for our contextualized soft token with one layer ($N=0$), the four tokens focus on different parts of the text. The last token focuses specifically on the meaning of <pad>, while the first three tokens have significantly higher attention values on the \textbf{key} part (\textbf{key}-value pair). This shows that the extracted information represents the index of information that our backbone LM should focus on. For contextualized soft tokens with two layers ($N=1$), we observe variations across the four attention distributions as well. Though the first token still focuses on the missing information, the rest of the tokens focus on the encoded meaningful \textbf{values} (key-\textbf{value} pairs). This shows that for $N>0$, the extracted soft token primarily represents the content of entities, which can be directly fed to our backbone LM.

\subsection{Results for the Information Augmentation Module. (I3)}

\stitle{\bl{Main Results.}} \bl{We conduct experiments on augmenting information using LLMs. Considering LLM's ability to be directly used to GEM, we also compare fine-tuned LLMs, zero-shot/few-shot inference for LLMs, and using LLMs as a labeler.}

\bl{For LLM's information augmentation, we augment information with GPT3.5 and conduct the experiments on four datasets (REL-TEXT, S-T-W, S-T-C, and WDC) that miss information more severely. As we describe in Sec.~\ref{sec:llmenhance}, for datasets with more homogeneity (REL-TEXT, WDC), we augment information with fixed attribute keys. For REL-TEXT (academic), we augment with \{title, abstract\}. For WDC (product), we augment with \{capacity, color, frequency, keywords, language, model number, product identifier, release year, resolution, size, speed, weight\}. For datasets containing heterogeneous data (S-T-W, S-T-C), we follow the augmentation at the granularity of data instances. We present the results of our model with augmented information in Tab.~\ref{tab:my-table}.}

\bl{For fine-tuned LLMs, we include the results of fine-tuned GPT3. In Tab.~\ref{tab:my-table}, GPT3 (Fine-tune) represents fine-tuning with structured input, while GPT3 (Fine-tune NL) represents fine-tuning with the natural language serialization.}

\bl{For zero-shot/ few-shot LLM inference, we list the zero-shot and few-shot ($2$, $5$ shot) results of GPT3.5 in Tab.~\ref{tab:my-table}. For zero-shot inference, we concatenate the two entities with prompts and query the LLM for an answer. For few-shot ($k$-shot) inference, we retrieve $k$ positive and $k$ negative query pairs in the labeled training pairs randomly or according to TF-IDF statistics. The pairs and labels are provided before the query prompts as in-context samples. The prompts for zero/few-shot are the same as previous work ~\cite{peeters2023using}. }

\bl{For using LLM as a labeler, we use a fine-tuned GPT3 model to provide labels for the $90\%$ unused training samples. The results are shown in Tab.~\ref{tab:my-table}, where $ax$ labels means we added $a$ times of the original training labels to the training set. }

\bl{As we can see for Tab.~\ref{tab:my-table}, augmenting information constantly boosts the performance of our base model. Though directly fine-tuning GPT-3 is promising on some datasets, our model can achieve comparable results compared with LLMs with the augmented information from LLMs. For datasets where LLMs achieve a good result (WDC), the improvement brought by the augmented information is also larger, which is quite intuitive. For the zero-shot/ few-shot results, we find out that fine-tuning is necessary even for LLMs for noisy GEM tasks. For the labeling experiment, we illustrate that training our model with augment information achieves better results than training our model with LLM provided labels (up to 3x labels than the original training data).}
\begin{figure*}[]
\centering
\begin{tabular}[t]{cc}

\subfigure[\bl{F1 results augmenting with varying uncertainty score thresholds}  ]{  
\hspace{1cm}
      \psfig{figure=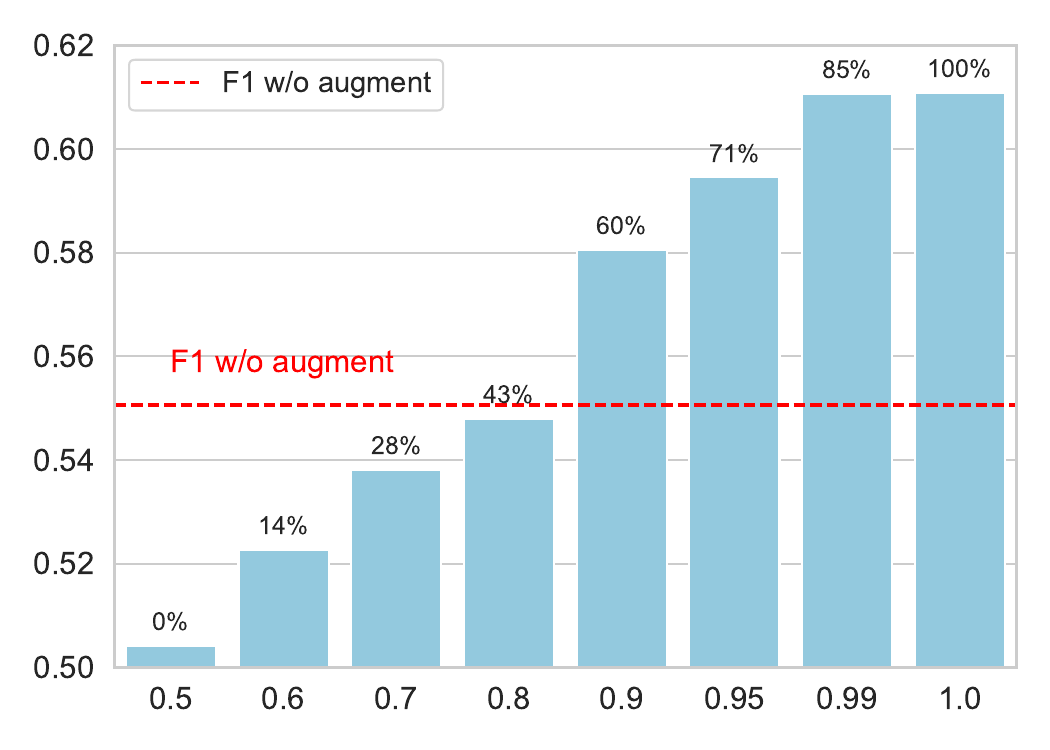,width=0.34
\linewidth }
      \label{fig:dblp_k}
      }
\hspace{1cm}
\subfigure[Case study for information augmentation on WDC ]{
      \psfig{figure=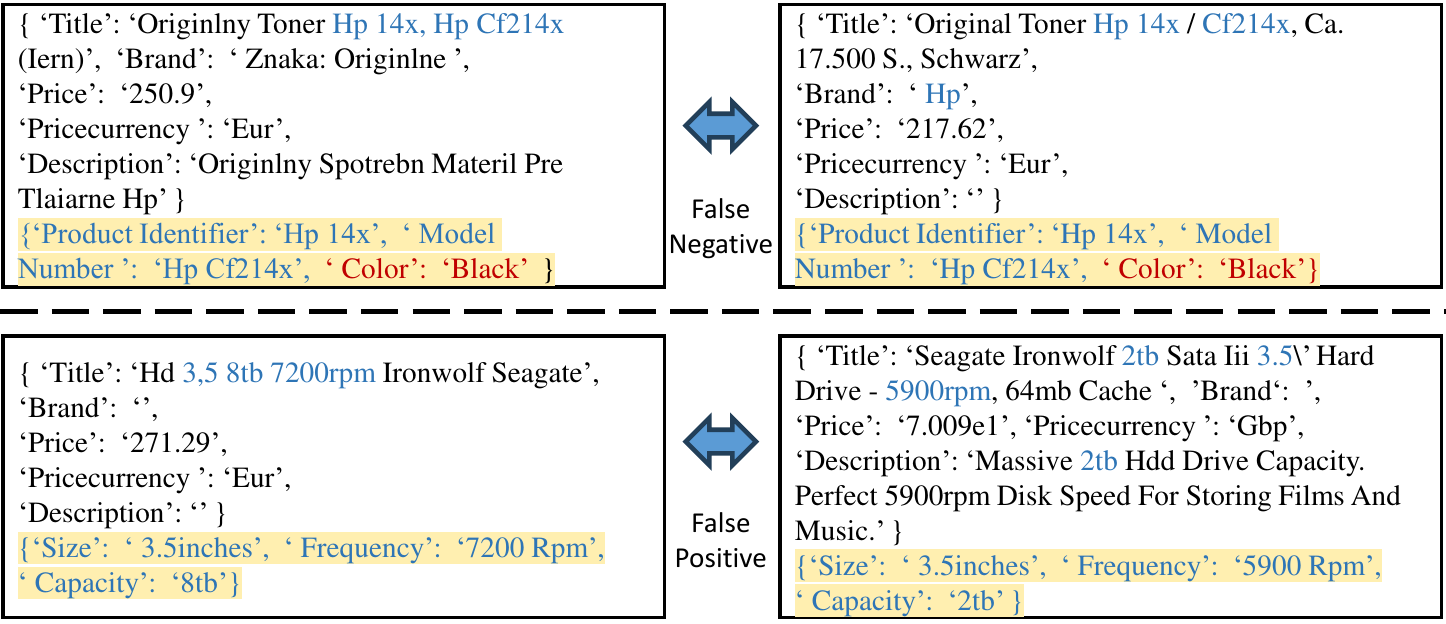,width=0.55\linewidth }
      \label{fig:dblp_k}
      }
\end{tabular}
%\vspace{-0.4cm}
\caption{\bl{Effectiveness of the uncertainty-based strategy and case study for information augmentation on the WDC dataset.} }
\label{fig:casestudy}
%\vspace{-0.5cm}
\end{figure*} 

\stitle{Effectiveness of the Uncertainty-based Information Augmentation Strategy.} We show the effectiveness of our augmenting strategy for the WDC dataset in Fig.~\ref{fig:casestudy} (a). We use $H (P(y|(e,e_i)))=max(P(y=0|(e,e_i)),P(y=1|(e,e_i))$ in this experiment. The x-axis represents the threshold of uncertainty scores, the y-axis represents the F1 results, the number on the bar represents the percentage of supplementary entities among all tested entities, and the red line represents APrompt4EM without information augmentation. As we can see, though augmenting part of the test data harms the consistency of the distribution between training and test data, augmenting half of the whole entity can bring comparable results with the base model. Augmenting $85\%$ of the whole entities can achieve similar results as augmenting all entities, as many comparing pairs are not confusing.

\begin{table}[]
\centering

\caption{F1 results of methods considering using LLMs.}
\label{tab:my-table}
\resizebox{0.45\textwidth}{!}{%
\begin{tabular}{l|llll}
\hline
           & REL-TEXT & S-T-W & S-T-C & WDC   \\ \hline
GPT3 (Fine-tune) & 64.8 & 6.3 & \textbf{85.7} & \textbf{67.5}  \\
GPT3 (Fine-tune NL) & \textbf{67.5} & 5.5 & 84.7 & 65.9   \\ \hline

\bl{GPT3.5 (0-shot)} & \bl{7.3} & \bl{9.0} & \bl{45.9} & \bl{22.1}  \\
\bl{GPT3.5 (2-shot,TF-IDF)} & \bl{46.6} & \bl{18.7} & \bl{31.8} & \bl{32.1}\\
\bl{GPT3.5 (2-shot,random)} & \bl{37.9} & \bl{21.6} & \bl{37.7} & \bl{45.4}
  \\ 
\bl{GPT3.5 (5-shot,TF-IDF)} & \bl{55.4} & \bl{16.4} & \bl{15.8} & \bl{33.6}  \\
\bl{GPT3.5 (5-shot,random)} & \bl{39.5} & \bl{20.8} & \bl{36.7} & \bl{39.9}  \\
\hline
Ours & 64.7 & 62.2 & 78.6 & 55.1 \\
\bl{Ours+1x Label}  & \bl{64.5} & \bl{49.4} & \bl{79.9} & \bl{55.3} \\
\bl{Ours+2x Label}& \bl{64.5} & \bl{40.6} & \bl{81.8} & \bl{58.4} \\
\bl{Ours+3x Label}& \bl{64.2} & \bl{29.3} & \bl{82.2} & \bl{58.8} \\ 
Ours+ aug. & \textbf{67.5} & \textbf{62.4} & 81.2 & 61.2 \\ \hline
\end{tabular}%
} 
\end{table}

\begin{figure}[]
\centering
\begin{tabular}[t]{cc}
\subfigure[Cost of the original dataset  ]{
      \psfig{figure=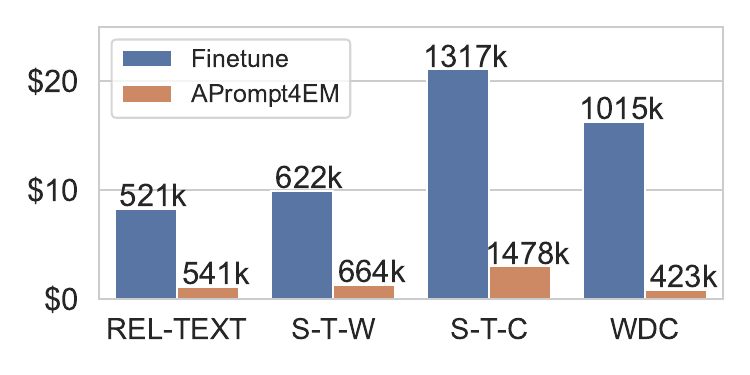,width=.49\linewidth }
      \label{fig:dblp_k}
      }
\subfigure[Cost of the blocking dataset ]{
      \psfig{figure=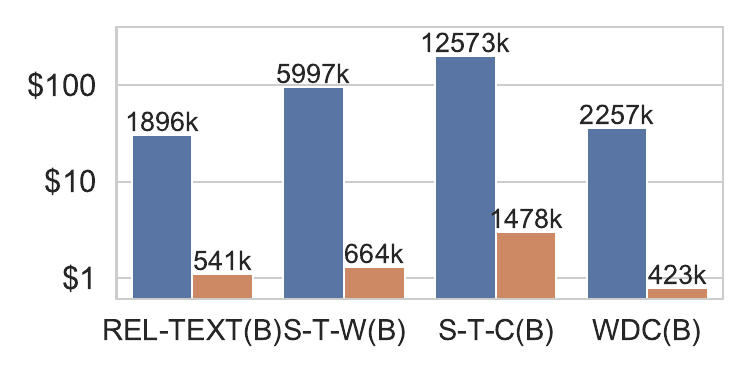,width=.49\linewidth }
      \label{fig:twitter_k}
      }
\end{tabular}
%\vspace{-0.4cm}
\caption{Cost analysis of our method: cost comparison between our information augmentation and direct inference.} 
\label{fig:costeff}
%\vspace{-0.5cm}
\end{figure}

\stitle{Cost Analysis of Our Method.}
We present the cost of our economical information augmentation method and the fine-tuning method using GPT3.5 in Fig.~\ref{fig:costeff}. We conduct two experiments on the original labeled dataset and a synthetic blocking dataset. For the blocking dataset, a simple TF-IDF blocker is used and a threshold is set to make each entity have an average of five candidates to be compared. We present the approximate token numbers and the total cost of the two tasks. The approximate token numbers are presented as numbers above the bars, and the cost is represented as the height of the bar. According to the official OpenAI pricing, the inference cost of a personal fine-tuned model is $0.016\$/1k$ tokens, while the inference cost of the original model is $0.002\$/1k$ tokens. Hence, the cost of inference using a fine-tuned model is much higher. As we can see, for the partly labeled original dataset, such as REL-TEXT, S-T-W, and S-T-C, the pairs tested are primarily from different entities, so our augmentation strategy benefits mainly from the pricing difference. In WDC, where the labeled samples contain duplicate entities, our strategy uses only half of the tokens. For scenarios closer to real-world matching applications, our augmentation method calls for $10\%$-$40\%$ tokens, as established in Sec.~\ref{sec:llmenhance} and Fig.~\ref{fig:costeff}, resulting in an even lower total cost.

\stitle{Case Study.} We present a case study of the WDC dataset. Two initially false negative and false positive pairs are presented in Fig.~\ref{fig:casestudy} (b). After information augmentation, both pairs are correctly classified. The augmentation information of the first two entities neglects the redundant description information of the first entity and points out that the key information of the two entities is the same. For the second example, though the same `ironwolf seagate 3.5' information confuses the LM, the augmented information successfully captures the two entities' difference in frequency and capacity in the redundant information, which helps our backbone LM change its decision. \bl{Besides, the information marked in red can be extracted from the original description, while the information marked in blue is somehow `recalled' by the LLM, which demonstrates the strong ability of the LLM to supplement information.  }

\section{Related Work}
\label{sec:related}

EM, as a crucial and challenging task in data integration, has received much attention in the past decades. Early works focus on rule-based methods~\cite{elmagarmid2014nadeef, wang2011entity,bilenko2003adaptive,kandhan2010sigmatch}, active sampling in EM~\cite{bellare2012active,meduri2020comprehensive}, and human labeling for EM~\cite{gokhale2014corleone,wang2012crowder}. In 2017-2018, DeepMatcher~\cite{mudgal2018deep} and DeepER~\cite{ebraheem2017deeper} first introduce deep learning into EM. 

\stitle{\bl{Fine-tuning methods for EM.}} \bl{After 2017, pre-trained language models (PLMs), which have powerful semantic expression abilities, have shown the capability of performing multiple NLP tasks~\cite{liu2023pre}, including the EM task. In the groundbreaking work of adopting PLMs on the EM task, Ditto~\cite{li2020deep}, the EM task is interpreted as a sequence classification problem. A structured entity $e$ is first serialized into a language sequence $serialize(e)$. The serialized sequences for two entity pairs are concatenated with two special separation tokens, $[CLS]$ and $[SEP]$, and the sequence is input to a PLM (i.e., BERT~\cite{devlin2018bert},  RoBERTa~\cite{liu2019roberta}, DistilBERT~\cite{sanh2019distilbert}). A task-specific   classification layer is added after the output of the $[CLS]$ token, and the model is tuned on the training set until convergence with its parameters initialized from the pre-trained model. }

\stitle{\bl{Prompt-tuning for EM.}} \bl{Different from the pre-train, fine-tune training paradigm, which trains a model for predicting the label distribution of an input and its label, the prompt-based tuning paradigm directly models the probability of an input text and computes the label probability referring to the text probability~\cite{liu2023pre}.}

\bl{The prompt templates used in prompt tuning can be categorized into discrete prompts and continuous prompts~\cite{liu2023pre}. For discrete prompts, apart from the manual design method, mining~\cite{shin2020autoprompt}, paraphrasing~\cite{jiang2020can}, and generation methods~\cite{gao2020making,guo2021text} are proposed to discover better prompts for downstream tasks. As for continuous prompts, representative methods include prefix tuning~\cite{li2021prefix}, Optiprompt~\cite{zhong2021factual}, and P-tuning~\cite{liu2023gpt}. Prefix tuning proposes adding a continuous prefix before the prompt content and tuning it during training, while the other parts of the model (main prompt, LM) are frozen. Optiprompt proposes adding soft tokens initialized by discrete prompts. As for P-tuning, trainable variables are introduced to the prompts  through a BiLSTM model.}

\bl{To address the low-resource challenge, PromptEM~\cite{wang2022promptem} first adopts the P-tuning approach to solve the EM problem. Our work, addressing the low-resource GEM task, is also based on the prompt tuning paradigm because it is proven effective in low-resource situations. It is worth noting that previous works tune one prompt for all data instances in a task, which is different from our contextualized prompt for each data instance. }

\stitle{\bl{Other perspectives for EM.}} \bl{Many works aim at the difficulty of the EM task in different complex real-world scenarios. Machamp~\cite{wang2021machamp} first presents the concept and a benchmark of GEM, and Machop~\cite{wang2022machop} provides a new framework for GEM considering the alignment at the attribute granularity. FlexER~\cite{genossar2023flexer} focuses on matching entities represented with multiple intents. ZeroER~\cite{wu2020zeroer} and TDMatch~\cite{ahmadi2022unsupervised} focus on more extreme matching scenarios where no supervision is provided.} 

\bl{Many other works bring in novel techniques to boost EM performance. A series of works~\cite{loster2021knowledge,zhao2019auto,tu2022domain} focus on domain adaptation and knowledge transferring in EM. Another series of works~\cite{miao2021rotom,wang2023sudowoodo,peeters2022supervised} discuss data augmentation in EM: Rotom~\cite{miao2021rotom} aims at generating augmented data for data integration; Sudowoodo~\cite{wang2023sudowoodo} and Supcon-EM~\cite{peeters2022supervised} focus on self-contrastive pre-training embeddings for EM. Recently, the usage of foundation models in EM are discussed in ~\cite{narayan2022can,peeters2023using}. }

\section{Conclusion}
\label{sec:conclusion}
In this paper, we propose APrompt4EM for low-resource GEM, in which the contextualized soft token prompt in natural language form helps our model extract critical information from noisy information, and information augmentation from LLMs helps bridge the semantic gap. Through extensive experiments and case studies, we report the superior performance of our model and illustrate the importance of all the techniques that we use in the low-resource GEM task. We report that the augmented model can achieve comparable results with fine-tuned LLMs using fewer API fees. We plan to further explore the possibility of more automatic adding and compressing for prompts in EM, e.g., via reinforcement learning.

\clearpage

\bibliographystyle{ACM-Reference-Format}
\bibliography{main}
 
% Please add the following required packages to your document preamble:
% \usepackage{graphicx}

\end{document}